\documentclass[letterpaper,twocolumn,10pt]{article}
\usepackage{usenix-2020-09}

\usepackage{graphicx}
\usepackage{subcaption}
\usepackage{booktabs}
\usepackage{multirow}
\usepackage[utf8]{inputenc}

\makeatletter
  \g@addto@macro{\UrlBreaks}{\UrlOrds}
\makeatother

\usepackage[inline]{enumitem}
\setlist[itemize]{leftmargin=*}
\usepackage{inconsolata}

\usepackage{amsmath}
\usepackage{amssymb}
\usepackage{tabularx}

\ifdefined\OverleafCompile
  \usepackage[newfloat=true,frozencache=true]{minted}
\else
  \usepackage[newfloat=true,frozencache=true]{minted}
\fi

\ifdefined\HabitatProceedings
\fi

\newcommand{\thetool}{Habitat}
\newcommand{\kernelsim}{wave scaling}
\newcommand{\Kernelsim}{Wave scaling}

\newcommand{\FLOPSMaxError}{64.9\%}
\newcommand{\FLOPSAtLeastError}{42.5\%}
\newcommand{\HeuristicsAvgError}{10.2\%}
\newcommand{\HeuristicsMaxError}{21.8\%}

\newcommand{\EteAvgError}{11.8\%}
\newcommand{\ResNetAvgError}{13.4\%}
\newcommand{\TransformerAvgError}{12.6\%}
\newcommand{\GNMTAvgError}{11.2\%}
\newcommand{\InceptionAvgError}{9.5\%}
\newcommand{\DCGANAvgError}{12.3\%}

\newcommand{\MLPOpAvgError}{18.0\%}
\newcommand{\WaveOpAvgError}{29.8\%}

\newcommand{\CloudCaseAvgError}{10.7\%}
\newcommand{\GeneralCaseAvgError}{7.7\%}
\newcommand{\GeneralCaseMaxSpeedup}{$1.1\times$}

\newcommand{\MixedPrecisionAvgError}{16.1\%}
\newcommand{\MixedPrecisionDaydreamOnlyAvgError}{10.7\%}

\newcommand{\MLPOperationCountProportion}{5\%}
\newcommand{\WaveOperationCountProportion}{95\%}
\newcommand{\MLPTimeProportion}{54\%}
\newcommand{\WaveTimeProportion}{46\%}

\newcommand{\TrainBatchSize}{512}
\newcommand{\TrainEpochs}{80}
\newcommand{\TrainLR}{$5 \times 10^{-4}$}
\newcommand{\TrainWeightDecay}{$10^{-4}$}
\newcommand{\TrainLRSecond}{$10^{-4}$}
\newcommand{\MLPNumLayers}{eight}
\newcommand{\MLPLayerSize}{1024}
\newcommand{\TrainLRAdjustAfter}{40}

\newcommand{\HabitatCodeLink}{\href{https://github.com/geoffxy/habitat}{github.com/geoffxy/habitat}}

\usepackage{titlesec}
\newcommand{\custombeforespacing}{0.5em plus 0.1em minus 0.1em}

\titlespacing*{\paragraph}{0em}{\custombeforespacing}{0.5em}

\begin{document}

\date{}
\title{A Runtime-Based Computational Performance Predictor for\\Deep
  Neural Network Training}

\author{
{\rm Geoffrey X.\ Yu}\\
University of Toronto\\Vector Institute
\and
{\rm Yubo Gao}\\
University of Toronto
\and
{\rm Pavel Golikov}\\
University of Toronto\\Vector Institute
\and
{\rm Gennady Pekhimenko}\\
University of Toronto\\Vector Institute
}

\maketitle

\begin{abstract}
Deep learning researchers and practitioners usually leverage GPUs
to help train their deep neural networks (DNNs) faster.
However, choosing \emph{which} GPU to use is challenging both because
\begin{enumerate*}[label=(\roman*)]
  \item there are many options, and
  \item users grapple with competing concerns: maximizing compute performance
    while minimizing costs.
\end{enumerate*}
In this work, we present a new practical technique to help users make informed
and cost-efficient GPU selections: make performance \emph{predictions} with
the help of a GPU that the user already has.
Our technique exploits the observation that, because DNN training consists of
repetitive compute steps, predicting the execution time of a single iteration
is usually enough to characterize the performance of an entire training
process.
We make predictions by scaling the execution time of each operation in a
training iteration from one GPU to another using either
\begin{enumerate*}[label=(\roman*)]
  \item wave scaling, a technique based on a GPU's execution model, or
  \item pre-trained multilayer perceptrons.
\end{enumerate*}
We implement our technique into a Python library called \thetool{} and find
that it makes accurate iteration execution time predictions (with an average
error of \EteAvgError{}) on ResNet-50, Inception v3, the Transformer, GNMT, and
DCGAN across six different GPU architectures.
\thetool{} supports PyTorch, is easy to use, and is open
source.\footnote{\thetool{} is available on GitHub:
  \HabitatCodeLink{}~\cite{habitat-code-yu21,habitat-models-yu21}}
\end{abstract}

\section{Introduction}\label{sec:introduction}
Over the past decade, deep neural networks (DNNs) have seen incredible success
across many machine learning
tasks~\cite{resnet-he16,vgg-simonyan15,alexnet-krizhevsky12,densenet-huang17,nmt-sutskever14,attention-vaswani17,bert-devlin19}---leading
them to become widely used throughout academia and industry.
However, despite their popularity, DNNs are not always straightforward to use
in practice because they can be extremely computationally-expensive to
train~\cite{mlperf,tbd-zhu18,dawn-coleman17,mlenergy-strubell19}.
This is why, over the past few years, there has been a significant and ongoing
effort to bring \emph{hardware acceleration} to DNN
training~\cite{cuda,a100,tpu-jouppi17,cerebras,graphcore,habana,trainium}.

As a result of this effort, today there is a vast array of hardware options for
deep learning users to choose from for training.
These options range from desktop and server-class GPUs (e.g.,
2080Ti~\cite{2080ti} and A100~\cite{a100}) all the way to specialized
accelerators such as the TPU~\cite{tpu-jouppi17}, AWS Trainium~\cite{trainium},
Gaudi~\cite{habana}, IPU~\cite{graphcore}, and Cerebras WSE~\cite{cerebras}.
Having all these options offers flexibility to users, but at the same time can
also lead to a paradox of choice: which hardware option should a researcher or
practitioner use to train their DNNs?

A natural way to start answering this question is to first consider
CUDA-enabled GPUs. This is because they
\begin{enumerate*}[label=(\roman*)]
  \item are commonly used in deep learning;
  \item are supported by all major deep learning software frameworks
    (PyTorch~\cite{pytorch-paszke19}, TensorFlow~\cite{tensorflow-abadi16}, and
    MXNet~\cite{mxnet-chen15});
  \item have mature tooling support (e.g., CUPTI~\cite{cupti}); and
  \item are readily available for rent \emph{and} purchase.
\end{enumerate*}
In particular, when considering GPUs, we find that that there are many
situations where a deep learning user needs to \emph{choose} a specific GPU to
use for training:
\begin{itemize}
  \item \textbf{Choosing between different hardware tiers.}
    In both academia and industry, deep learning users often have access to
    several \emph{tiers} of hardware:
    \begin{enumerate*}[label=(\roman*)]
      \item a workstation with a GPU used for development (e.g., 2080Ti),
      \item a private GPU cluster that is shared within their organization
        (e.g., RTX6000~\cite{rtx6000}), and
      \item GPUs that they can rent in the cloud (e.g., V100~\cite{v100}).
    \end{enumerate*}
    Each tier offers a different \emph{cost}, \emph{availability}, and
    \emph{performance} trade-off. For example, a private cluster might be
    ``free'' (in monetary cost) to use, but jobs may be queued because the
    cluster is also shared among other users. In contrast, cloud GPUs can
    be rented on-demand for exclusive use.

  \item \textbf{Deciding on which GPU to rent or purchase.}
    Cloud providers make many different GPUs available for rent (e.g.,
    P100~\cite{p100}, V100, T4~\cite{t4}, and A100~\cite{a100}), each with
    different performance at different prices. Similarly, a wide variety of
    GPUs are available for purchase (e.g., 2080Ti, 3090~\cite{3090}) both
    individually and as a part of pre-built workstations~\cite{lambdalabs}.
    These GPUs can vary up to $6\times$ in price~\cite{pricediff} and $6\times$
    in peak performance~\cite{nvidia-a100-benchmark}.

  \item \textbf{Determining how to schedule a job in a heterogeneous GPU
      cluster.} A compute cluster (e.g., operated by a cloud
    provider~\cite{sagemaker,vertexai,azureml}) may have multiple types of
    GPUs available that can handle a training workload. Deciding which GPU to
    use for a job will typically depend on the job's priority and performance
    on the GPU being
    considered~\cite{heterodlsched-narayanan20,gandivafair-chaudhary20}.

  \item \textbf{Selecting alternative GPUs.} When a desired GPU is unavailable
    (e.g., due to capacity constraints in the cloud), a user may want to select
    a different GPU with a comparable cost-normalized performance. For example,
    when training ResNet-50~\cite{resnet-he16} on Google Cloud~\cite{gcp-gpus},
    we find that both the P100 and V100 have similar cost-normalized
    throughputs (differing by just 0.8\%). If the V100 were to be
    unavailable,\footnote{In our experience, we often ran into situations where
      the V100 was unavailable for rent because the cloud provider had an
      insufficient supply.} a user may decide to use the P100 instead since the
    total training cost would be similar.
\end{itemize}
What makes these situations interesting is that there is not necessarily a
\emph{single} ``correct'' choice.
Users make GPU selections based on whether the performance benefits of the
chosen configuration are \emph{worth} the cost to train their DNNs.
But making these selections in an informed way is not easy, as performance
depends on many factors simultaneously:
\begin{enumerate*}[label=(\roman*)]
  \item the DNN being considered,
  \item the GPU being used, and
  \item the underlying software libraries used during training (e.g.,
    cuDNN~\cite{cudnn}, cuBLAS~\cite{cublas}).
\end{enumerate*}

To do this performance analysis today, the common wisdom is to either
\begin{enumerate*}[label=(\roman*)]
  \item directly measure the computational performance (e.g., throughput) by
    actually running the training job on the GPU, or
  \item consult existing benchmarks (e.g., MLPerf~\cite{mlperf}) published by
    the community to get a ``ballpark estimate.''
\end{enumerate*}
While convenient, these approaches also have their own limitations.
Making measurements requires users to already have access to the GPUs they are
considering; this may not be the case if a user is deciding whether or not to
\emph{buy} or \emph{rent} that GPU in the first place.
Secondly, benchmarks are usually only available for a subset of GPUs (e.g., the
V100 and T4) and only for common ``benchmark'' models (e.g.,
ResNet-50~\cite{resnet-he16} and the Transformer~\cite{attention-vaswani17}).
They are not as helpful if you need an accurate estimate of the performance of
a \emph{custom} DNN on a specific GPU (a common scenario when doing deep
learning research).

In this work, we make the case for a third complementary approach: making
performance \emph{predictions}.
Although predicting the performance of general compute workloads can be
prohibitively difficult due to the large number of possible program phases, we
observe that DNN training workloads are special because they contain
\emph{repetitive computation}.
DNN training consists of repetitions of the same (relatively short) training
iteration, which means that the performance of an entire training process can
be characterized by just a few training iterations.

\begin{listing}[t]
\begin{minted}[fontsize=\footnotesize]{python}
import habitat

tracker = habitat.OperationTracker(
  origin_device=habitat.Device.RTX2070,
)

with tracker.track():
    run_my_training_iteration()

trace = tracker.get_tracked_trace()
print("Pred. iter. exec. time: {:.2f} ms".format(
  trace.to_device(habitat.Device.V100).run_time_ms,
))
\end{minted}
\vspace{-1em}
\caption{An example of how \thetool{} can be used to make iteration execution
  time predictions.}
\label{lst:habitat-api}
\vspace{-0.2em}
\end{listing}

We leverage this observation to build a new technique that \emph{predicts} a
DNN's training iteration execution time for a given batch size and GPU using
both \emph{runtime information} and \emph{hardware characteristics}.
We make predictions in two steps:
\begin{enumerate*}[label=(\roman*)]
  \item we measure the execution time of a training iteration on an
    \emph{existing} GPU, and then
  \item we scale the measured execution times of each individual operation onto
    a \emph{different} GPU using either \kernelsim{} or pre-trained multilayer
    perceptrons (MLPs)~\cite{dlbook-goodfellow16}.
\end{enumerate*}
\Kernelsim{} is a technique that applies \emph{scaling factors} to the GPU
kernels in an operation, based on a mix of the ratios between the two GPUs'
memory bandwidth and compute units.
We use MLPs for certain operations (e.g., convolution) where the kernels used
differ between the two GPUs; we describe this phenomenon and the MLPs in more
detail in Sections~\ref{sec:overview} and \ref{sec:mlp-predictors}.
We believe that using an existing GPU to make operation execution time
predictions for a different GPU is reasonable because deep learning users often
already have a local GPU that they use for development.

We implement our technique into a Python library that we call \thetool{}, and
evaluate its prediction accuracy on five DNNs that have applications in image
classification, machine translation, and image generation:
\begin{enumerate*}[label=(\roman*)]
  \item ResNet-50,
  \item Inception v3~\cite{inception-szegedy15}
  \item the Transformer,
  \item GNMT~\cite{gnmt-wu16}, and
  \item DCGAN~\cite{dcgan-radford16}.
\end{enumerate*}
We use \thetool{} to make iteration execution time predictions across six
different GPUs and find that it makes accurate predictions with an average
error of \EteAvgError{}.
Additionally, we present two case studies to show how \thetool{} can be used to
help users make accurate cost-efficient GPU selections according to their needs
(Section~\ref{sec:case-studies}).

We designed \thetool{} to be easy and practical to use
(see Listing~\ref{lst:habitat-api}).
\thetool{} currently supports PyTorch~\cite{pytorch-paszke19} and is open
source: \HabitatCodeLink{}~\cite{habitat-code-yu21,habitat-models-yu21}.

In summary, this work makes the following contributions:

\begin{itemize}
  \item \Kernelsim{}: a new technique that scales the execution time of a
    kernel measured on one GPU to a different GPU by using scaled ratios
    between the
    \begin{enumerate*}[label=(\roman*)]
      \item number of compute units on each GPU, and
      \item their memory bandwidths.
    \end{enumerate*}

  \item The implementation and evaluation of \thetool{}: a new library that
    uses \kernelsim{} along with pre-trained MLPs to predict the execution time
    of DNN training iterations on different GPUs.
\end{itemize}

\section{Why Predict the Computational Training Performance of DNNs on
  Different GPUs?}
This work presents a new practical technique for predicting the execution time
of a DNN training iteration on different GPUs, with the goal of helping deep
learning users make informed cost-efficient GPU selections.
However, a common first question is to ask why we need to make these
performance predictions in the first place.
Could other performance comparison approaches (e.g., simple heuristics or
measurements) be used instead?
In this section, after providing some background about DNN training, we outline
the problems with these alternative approaches to further motivate the need for
practical performance predictions.

\subsection{Background on DNN Training}\label{sec:background-dnn}
DNNs, at their heart, are mathematical functions that produce predictions given
an input and a set of \emph{learned} parameters, also known as
\emph{weights}~\cite{dlbook-goodfellow16}. They are built by combining together
a series of different \emph{layers}, each of which may contain weights. The
layers map to mathematical operations. For example, a fully connected layer is
implemented using matrix multiplication~\cite{dlbook-goodfellow16}. To produce
predictions, a DNN takes a tensor (an $n$-dimensional array) as input and
applies the operations associated with each layer in sequence.

\paragraph{Training.}
A DNN learns its weights in an \emph{iterative} process called training. Each
training iteration operates on a batch of labelled inputs and consists of a
\emph{forward} pass, \emph{backward} pass (using
backpropagation~\cite{backprop-rumelhart86}), and \emph{weight update}. The
forward and backward passes compute gradients for the weights, which are then
used by an optimization algorithm (e.g., stochastic gradient
descent~\cite{sgd-bottou10} or Adam~\cite{adam-kingmaba15}) to update the
weights so that the DNN produces better predictions. These steps are
\emph{repeated} until the DNN makes acceptably accurate predictions.

\paragraph{Computational performance.}
Although conceptually simple, prior work has shown that DNN training can be an
extremely time-consuming
process~\cite{mlperf,dawn-coleman17,tbd-zhu18,mlenergy-strubell19}. There are
two primary factors that influence the time it takes a DNN to reach an
acceptable accuracy during training~\cite{abm-mitliagkas16}: (i) statistical
efficiency, and (ii) hardware efficiency. Statistical efficiency governs the
\emph{number} of training iterations (i.e., weight updates) required to reach a
target test accuracy whereas hardware efficiency governs how \emph{quickly} a
training iteration runs. In this work, we focus on helping deep learning users
make informed cost-efficient hardware configuration selections to improve their
DNN's \emph{hardware efficiency}. As a result, we compare the performance of
different GPUs when training a DNN using the \emph{time it takes a training
  iteration to run}. This metric equivalently captures the training throughput
for that particular DNN.

\subsection{Why Not Measure Performance Directly?}
Perhaps the most straightforward approach to compare the performance of
different GPUs is to just measure the iteration execution time (and hence,
throughput) on each GPU when training a given DNN.
However, this approach also has a straightforward downside: it requires the
user to actually have access to the GPU(s) being considered in the first place.
If a user is looking to buy or rent a cost-efficient GPU, they would ideally
want to know its performance on their DNNs \emph{before} spending money to get
access to the GPU.

\begin{figure}[t]
  \includegraphics[width=\columnwidth]{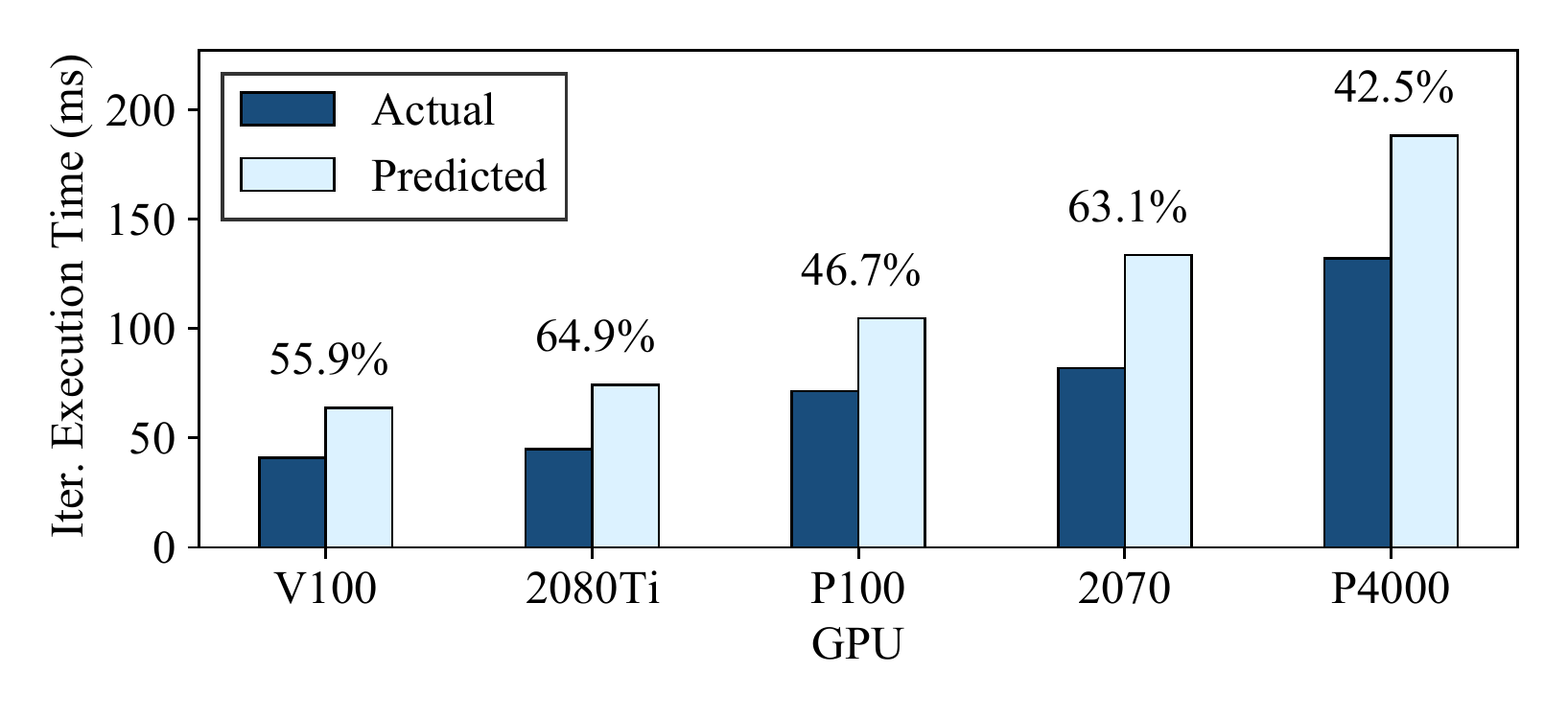}
  \vspace{-2.8em}
  \caption{DCGAN iteration execution time predictions, and their errors, made
    from the T4 using peak FLOPS ratios between the devices. Using simple
    heuristics can lead to high prediction errors.}
  \label{fig:heuristics-gflops}
\end{figure}

\subsection{Why Not Apply Heuristics?}
Another approach is to use heuristics based on the hardware specifications
published by the manufacturer.
For example, one could use the ratio between the peak floating point operations
per second (FLOPS) of two GPUs or the ratio between the number of CUDA cores on
each GPU.
The problem with this approach is that these heuristics do not always work.
Heuristics often assume that a DNN training workload can exhaust all the
computational resources on a GPU, which is not true in
general~\cite{tbd-zhu18}.

To show an example of when simple heuristics do not work well, we use a GPU's
peak FLOPS to make iteration execution time predictions.
We measure the execution time of a DCGAN training iteration on the
T4\footnote{We use a batch size of 128 LSUN~\cite{lsun-yu15} synthetic inputs.
  See Section~\ref{sec:methodology} for details about our methodology.} and
then use this measurement to predict the iteration execution time on
\emph{different} GPUs by multiplying by the ratio between the devices' peak
FLOPS.
Figure~\ref{fig:heuristics-gflops} shows the measured and predicted execution
times on each GPU, along with the prediction error as a percentage.
The main takeaway from this figure is that using simple heuristics can lead to
high prediction errors; the highest prediction error in this experiment is
\FLOPSMaxError{}, and all the prediction errors are at least
\FLOPSAtLeastError{}.
In contrast, \thetool{} can make these exact same predictions with an average
error of \HeuristicsAvgError{} (maximum \HeuristicsMaxError{}).

\subsection{Why Not Use Benchmarks?}
A third potential approach is to consult published benchmarking
results~\cite{mlperf,tbd-zhu18,dawn-coleman17,nvidia-benchmarks}.
However, the problem with relying on benchmarking results is that they are
limited to a set of ``common'' DNNs (e.g., ResNet-50 or BERT~\cite{bert-devlin19})
and are usually only available for a small selection of GPUs (e.g., the T4, V100,
and A100).
Moreover, benchmarking results also vary widely among different models and
GPUs~\cite{mlperf,tbd-zhu18,nvidia-benchmarks}.
Therefore if no results exist for the GPU(s) a user is considering, or if a
user is working with a new DNN architecture, there will be no benchmark results
for them to consult.

\subsection{Why Not Always Use the ``Best'' GPU?}
Finally, a fourth approach is to always use the most ``powerful'' GPU available
with the assumption that GPUs are already priced based on their performance.
Why make performance predictions when the cost-efficiency of popular GPUs
should be the same?
However, this assumption is a misconception.
Prior work has already shown examples where the performance benefits of
different GPUs changes depending on the
model~\cite{heterodlsched-narayanan20,gandivafair-chaudhary20,tbd-zhu18}.
In this work, we also show additional examples in our case studies
(Section~\ref{sec:case-studies}) where
\begin{enumerate*}[label=(\roman*)]
  \item cost-efficiency leads to selecting a different GPU, and
  \item where the V100 does not offer significant performance benefits over a
    common desktop-class GPU (the 2080Ti).
\end{enumerate*}

\paragraph{Summary.}
Straightforward approaches that users might consider to make GPU selections all
have their own downsides.
In particular, existing approaches either require access to the GPUs themselves
or are only applicable for common DNNs and GPUs.
Therefore there is a need for a complementary approach: making performance
predictions---something that we explore in this work.

\section{\thetool{}}
Our approach to performance predictions is powered by three key observations.
In this section, after describing these observations, we outline the key ideas
behind \thetool{}.

\subsection{Key Observations}
\paragraph{Observation 1: Repetitive computation.}
While training a DNN to an acceptable accuracy can take on the order of
hours to days~\cite{mlperf,dawn-coleman17,tbd-zhu18}, a single training
iteration takes on the order of hundreds of \emph{milliseconds}. This
observation improves the predictability of DNN training as we can characterize
the performance of an entire DNN training session using the performance of a
single iteration.

\paragraph{Observation 2: Common building blocks among DNNs.}
Although DNNs can consist of hundreds of operations, they are built using a
relatively \emph{small} set of unique operations. For example, convolutional
neural networks typically comprise convolutional, pooling, fully connected, and
batch normalization~\cite{batchnorm-ioffe15} layers. This observation reduces
the problem of predicting the performance of an arbitrary DNN's training
iteration to developing prediction mechanisms for a small set of operations.

\paragraph{Observation 3: Runtime information available.}
When developing DNNs, users often have a GPU available for use in their
workstations. These GPUs are used for development purposes and are not
necessarily chosen for the highest performance (e.g., 1080Ti~\cite{1080ti},
TITAN Xp~\cite{titanxp}). However, they can be used to provide valuable runtime
information about the GPU kernels that are used to implement a given DNN. In
Section~\ref{sec:wave-scaling}, we describe how we can leverage this runtime
information to predict the performance of the GPU kernels on different GPUs
(e.g., from a desktop-class GPU such as the 2080Ti~\cite{2080ti} to a
server-class GPU such as the V100~\cite{v100,volta-nvidia}).

\subsection{\thetool{} Overview}\label{sec:overview}
\thetool{} records information at runtime about a DNN training iteration for a
specific batch size on a given GPU \emph{(Observation 3)} and then uses that
information to predict the training iteration execution time on a
\emph{different} GPU (for the same batch size). Predicting the iteration
execution time is enough \emph{(Observation 1)} to compute metrics about the
entire training \emph{process} on different GPUs. These predicted metrics, such
as the training throughput and cost-normalized throughput, are then used by
end-users (e.g., deep learning researchers) to make informed hardware
selections.

To actually make these predictions for a different GPU, \thetool{} predicts the
new execution time of each individual operation in a training iteration.
\thetool{} then adds these predicted times together to arrive at an execution
time prediction for the entire iteration. For an individual operation,
\thetool{} makes predictions using either
\begin{enumerate*}[label=(\roman*)]
  \item \kernelsim{} (Section~\ref{sec:wave-scaling}), or
  \item pre-trained MLPs (Section~\ref{sec:mlp-predictors}).
\end{enumerate*}

The reason why \thetool{} uses two techniques together is that \kernelsim{}
assumes that the \emph{same} GPU kernels are used to implement a given DNN
operation on each GPU. However, some DNN operations are implemented using
\emph{different} GPU kernels on different GPUs (e.g., convolutions, recurrent
layers). This is done for performance reasons as these operations are typically
implemented using proprietary kernel libraries that leverage GPU
architecture-specific kernels (e.g., cuDNN~\cite{cudnn-chetlur14},
cuBLAS~\cite{cublas}). We refer to these operations as \emph{kernel-varying},
and scale their execution times to different GPUs using pre-trained MLPs.
\thetool{} uses \kernelsim{} for the rest of the operations, which we call
\emph{kernel-alike}.

\subsection{Wave Scaling}\label{sec:wave-scaling}
\Kernelsim{} works by scaling the execution times of the \emph{kernels} used to
implement a kernel-alike DNN operation. The computation performed by a GPU
kernel is partitioned into groups of threads called \emph{thread
  blocks}~\cite{fermi-nvidia09}, which typically execute in concurrent groups,
resulting in \emph{waves} of execution. The key idea behind \kernelsim{} is to
compute the number of \emph{thread block waves} in a kernel and scale the wave
execution time using \emph{ratios} between the origin and destination GPUs.

We describe \kernelsim{} formally in Equation~\ref{eqn:wave-scaling}. Let
$T_i$ represent the execution time of the kernel on GPU $i$, $B$ the number of
thread blocks in the kernel, $W_i$ the number of thread blocks in a wave on GPU
$i$, $D_i$ the memory bandwidth on GPU $i$, and $C_i$ the clock frequency on
GPU $i$. Here we let $i \in \{o, d\}$ to represent the origin and destination
GPUs. By measuring $T_o$ (\emph{Observation 3}), \kernelsim{} predicts $T_d$
using
\begin{equation}\label{eqn:wave-scaling}
  T_d =
  \left\lceil{\frac{B}{W_d}}\right\rceil
  \left(\frac{D_o}{D_d} \frac{W_d}{W_o}\right)^\gamma
  \left(\frac{C_o}{C_d}\right)^{1-\gamma}
  \left\lceil{\frac{B}{W_o}}\right\rceil^{-1}
  T_o
\end{equation}
where $\gamma \in [0, 1]$ represents the ``memory bandwidth boundedness'' of
the kernel. \thetool{} selects $\gamma$ by measuring the kernel's arithmetic
intensity and then leveraging the roofline model~\cite{roofline-williams09}
(see Section~\ref{sec:selecting-gamma}).

As shown in Equation~\ref{eqn:wave-scaling}, \kernelsim{} uses the ratios
between the GPUs'
\begin{enumerate*}[label=(\roman*)]
  \item memory bandwidths,
  \item clock frequencies, and
  \item the size of a wave on each GPU.
\end{enumerate*}
The intuition behind factors (i) and (iii) is that a higher relative memory
bandwidth allows more memory requests to be served in parallel whereas having
more thread blocks in a wave results in more memory requests being made. Thus,
everything else held constant, waves in memory bandwidth bound kernels (i.e.,
large $\gamma$) should see speedups on GPUs with more memory bandwidth. The
intuition behind factor (ii) is that higher clock frequencies may benefit waves
in compute bound kernels (i.e., small $\gamma$).\footnote{The clock's impact on
  execution time depends on other factors too (e.g., the GPU's instruction set
  architecture). \Kernelsim{} aims to be a simple and understandable model and
  therefore does not model these complex effects.}

For large $\left\lceil{B/W_i}\right\rceil$ (i.e., when there are a large number
of waves) we get that $\left\lceil{B/W_i}\right\rceil \approx B/W_i$. In this
case, Equation~\ref{eqn:wave-scaling} simplifies to
\begin{equation}\label{eqn:wave-scaling-simple}
  T_d = \left(\frac{D_o}{D_d}\right)^\gamma
  \left(\frac{W_o}{W_d}\right)^{1-\gamma}
  \left(\frac{C_o}{C_d}\right)^{1-\gamma}
  T_o
\end{equation}
\thetool{} uses Equation~\ref{eqn:wave-scaling-simple} to predict kernel
execution times because we find that in practice, most kernels are composed of
many thread blocks.

\thetool{} computes $W_i$ for each kernel and GPU using the thread block
occupancy calculator that is provided as part of the CUDA Toolkit~\cite{cuda}.
We obtain $C_i$ from each GPU's specifications, and we obtain $D_i$ by
measuring the achieved bandwidth on each GPU ahead of time. Note that we make
these measurements once and then distribute them in a configuration file with
\thetool{}.

\subsection{MLP Predictors}\label{sec:mlp-predictors}
To handle kernel-varying operations, \thetool{} uses pre-trained MLPs to make
execution time predictions.
We treat this prediction task as a regression problem: given a series of input
features about the operation and a target GPU (described below), predict the
operation's execution time on that target GPU.
We learn an MLP for each kernel-varying operation that \thetool{} currently
supports:
\begin{enumerate*}[label=(\roman*)]
  \item convolutions (2-dimensional),
  \item LSTMs~\cite{lstm-hochreiter97},
  \item batched matrix multiplies, and
  \item linear layers (matrix multiply with an optional bias term).
\end{enumerate*}
As we show in Section~\ref{sec:evaluation}, relatively few DNN operations are
kernel-varying\footnote{This is, in part, because implementing performant
  architecture-specific kernels for each kernel-varying operation takes
  significant engineering effort.} and so training separate MLPs for each of
these operations is a feasible approach.
Furthermore, these MLPs can be used for many different DNNs as these operations
are common ``building blocks'' used in DNNs (\emph{Observation 2}).

\paragraph{Input features.}
Each operation-specific MLP takes as input:
\begin{enumerate*}[label=(\roman*)]
  \item layer dimensions (e.g., the number of input and output
    channels in a convolution);
  \item the memory capacity and bandwidth on the target GPU;
  \item the number of streaming multiprocessors (SMs) on the
    target GPU; and
  \item the peak FLOPS of the target GPU, specified by the manufacturer.
\end{enumerate*}

\paragraph{Model architecture.}
Each MLP comprises an input layer, \MLPNumLayers{} hidden layers, and an output
layer that produces a single real number---the predicted execution time (this
includes the forward and backward pass) for the MLP's associated operation.
We use ReLU activation functions in each layer and we use \MLPLayerSize{} units
in each hidden layer.
We outline the details behind our datasets and how these MLPs are trained in
Section~\ref{sec:mlp-impl-data}.

\section{Implementation Details}
\thetool{} is built to work with PyTorch~\cite{pytorch-paszke19}. However, the
ideas behind \thetool{} are general and straightforward to implement in other
frameworks as well. \thetool{} performs its analysis using a DNN's computation
graph, which is also available in other frameworks (e.g.,
TensorFlow~\cite{tensorflow-abadi16} and MXNet~\cite{mxnet-chen15}).

\subsection{Extracting Runtime Metadata}\label{sec:execution-trace}
\thetool{} extracts runtime metadata in a training iteration by ``monkey
patching'' PyTorch operations with special wrappers. These wrappers allow
\thetool{} to intercept and keep track of all the operations that run in one
training iteration, as they are executed. As shown in
Listing~\ref{lst:habitat-api}, users explicitly indicate to \thetool{} when to
start and stop tracking the operations in a DNN by calling \texttt{track()}.

\paragraph{Execution time.}
To measure the execution time of each operation, \thetool{} re-runs each
operation independently with the same inputs as recorded when the operation was
intercepted. \thetool{} also measures the execution time associated with the
operation's backward pass, if applicable. The reason why \thetool{} makes
measurements by re-running the individual operations is that the operations
could be very short (in execution time). Thus, \thetool{} needs to run them
multiple times to make accurate measurements. \thetool{} uses CUDA
events~\cite{cuda-events} to make these timing measurements.

\paragraph{Kernel metadata and metrics.}
\thetool{} uses CUPTI~\cite{cupti} to record execution times and metrics (see
Section~\ref{sec:selecting-gamma}) for the kernels used to implement each
operation in the DNN. This information is used by \kernelsim{}.

\subsection{Selecting Gamma\texorpdfstring{ ($\gamma$)}{}}\label{sec:selecting-gamma}
Recall from Section~\ref{sec:wave-scaling} that \kernelsim{} scales its ratios
using $\gamma$, a factor that represents the ``memory bandwidth boundedness''
of a kernel. In this section, we describe in more detail how \thetool{}
automatically selects $\gamma$ for each kernel.

\begin{figure}
  \includegraphics[width=\columnwidth]{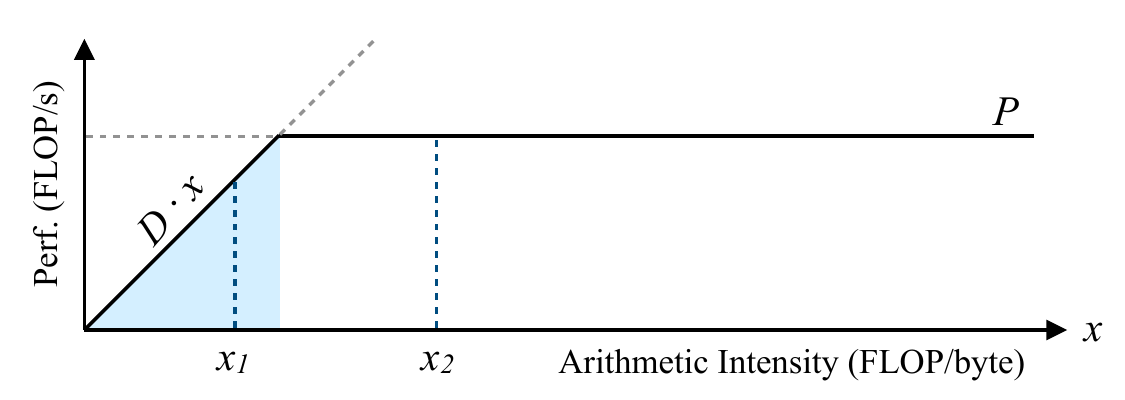}
  \vspace{-2em}
  \caption{An example roofline model. If a kernel's arithmetic intensity falls
    in the shaded region, it is considered memory bandwidth bound ($x_1$);
    otherwise, it is considered compute bound ($x_2$).}
  \label{fig:roofline-example}
\end{figure}

\paragraph{Roofline model.}
\thetool{} uses the roofline model~\cite{roofline-williams09} to estimate a
kernel's memory boundedness. Figure~\ref{fig:roofline-example} shows an
example roofline model.
The roofline model introduces the notion of a kernel's arithmetic intensity:
the number of floating point operations it performs per byte of data read or
written to memory (represented by $x$ in Figure~\ref{fig:roofline-example}).

A key idea behind the roofline model is that it models a kernel's peak
performance as the minimum of either the hardware's peak performance ($P$) or
the hardware's memory bandwidth times the kernel's arithmetic intensity ($D
\cdot x$)~\cite{roofline-williams09}. This minimum is shown by the solid line
in Figure~\ref{fig:roofline-example}.
The arithmetic intensity where these two limits meet is called the ``ridge
point'' ($R$), where $R = P / D$.
The model considers a kernel with an arithmetic intensity of $x$ to be
memory bandwidth bound if $x < R$ and compute bound otherwise.
For example, in Figure~\ref{fig:roofline-example}, a kernel with an arithmetic
intensity of $x_1$ would be considered memory bandwidth bound whereas a kernel
with an intensity of $x_2$ would be considered compute bound.

\Kernelsim{} leverages the observation that a kernel's arithmetic intensity is
fixed across GPUs (i.e., arithmetic intensity only depends on the kernel's
code). $R$ changes across GPUs because $P$ and $D$ vary among GPUs, but can be
computed using a GPU's performance specifications. Therefore, if \thetool{}
computes a kernel's arithmetic intensity, it can use the arithmetic intensity's
distance from the destination GPU's ridge point to estimate the kernel's memory
bandwidth boundedness (on the destination GPU).

\paragraph{Selecting $\gamma$.}
When profiling each kernel, \thetool{} gathers metrics that allow it to
empirically calculate the kernel's arithmetic intensity (floating point
efficiency, number of bytes read and written to DRAM). If we let $x$ be the
kernel's measured arithmetic intensity and $R = P/D$ for the destination GPU
(using the notation presented above), \thetool{} sets $\gamma$ using
\begin{equation}
  \gamma =
  \begin{cases}
    (-0.5/R) x + 1 & \mathrm{if\ } x < R \\
    0.5R/x       & \mathrm{otherwise}
  \end{cases}
\end{equation}
This equation means that $\gamma$ decreases linearly from 1 to 0.5 as $x$
increases toward $R$. After passing $R$, $\gamma$ approaches 0 as $x$
approaches infinity.

\paragraph{Practical optimizations.}
In practice, gathering metrics on GPUs is a slow process because the kernels
need to be replayed multiple times to capture all the needed performance
counters. To address this challenge, we make two optimizations:
\begin{enumerate*}[label=(\roman*)]
  \item we cache measured metrics, keyed by the kernel's name and its launch
    configuration (number of thread blocks and block size); and
  \item we only measure metrics for operations that contribute significantly to
    the training iteration's execution time (e.g., with execution times at or
    above the 99.5th percentile).
\end{enumerate*}
Consequently, when metrics are unavailable for a particular kernel, we set
$\gamma = 1$.
We believe that this is a reasonable approximation because kernel-alike
operations tend to be very simple (e.g., element-wise operations) and are
therefore usually memory bandwidth bound.

\subsection{MLPs: Data and Training}\label{sec:mlp-impl-data}
In this section, we describe the details behind \thetool{}'s MLPs: how we
\begin{enumerate*}[label=(\roman*)]
  \item collect training data,
  \item preprocess the data, and
  \item train the MLPs.
\end{enumerate*}

\subsubsection{Data Collection}
We gather training data by measuring the forward and backward pass execution
times of kernel-varying operations at randomly sampled \emph{input
  configurations}. An input configuration is a setting of an operation's
parameters (e.g., batch size and number of channels in a convolution). We use
predefined ranges for each operation's parameters, as described in more detail
below, and ignore any configurations that result in running out of memory.
We make these measurements for all six of the GPUs listed in
Section~\ref{sec:methodology}.
We use the same seed when sampling on different GPUs to ensure we have
measurements for the same random input configurations across all the GPUs. We
create the final dataset by joining data entries that have the same operation
and configuration, but with different GPUs.

\paragraph{2D convolutions.}
For convolutions, we vary the
\begin{enumerate*}[label=(\roman*)]
  \item batch size (1 -- 64),
  \item number of input (3 -- 2048) and output channels (16 -- 2048),
  \item kernel size (1 -- 11),
  \item padding (0 -- 3),
  \item stride (1 -- 4),
  \item image size (1 -- 256), and
  \item whether or not there is a bias weight.
\end{enumerate*}
We only sample configurations with square images and kernel sizes. During
sampling, we ignore any configurations that result in invalid arguments (e.g.,
a kernel size larger than the image). We selected these parameter ranges by
surveying the convolutional neural networks included in PyTorch's
\texttt{torchvision} package~\cite{torchvision}.

\paragraph{LSTMs.}
For LSTMs, we vary the
\begin{enumerate*}[label=(\roman*)]
  \item batch size (1 -- 128),
  \item number of input features (1 -- 1280),
  \item number of hidden features (1 -- 1280),
  \item sequence length (1 -- 64),
  \item number of stacked layers (1 -- 6),
  \item whether or not the LSTM is bidirectional, and
  \item whether or not there is a bias weight.
\end{enumerate*}

\paragraph{Batched matrix multiply (bmm).}
For a batched matrix multiply of $A \times B$ where $A \in \mathbb{R}^{n \times
  l \times m}$ and $B \in \mathbb{R}^{n \times m \times r}$, we vary the
\begin{enumerate*}[label=(\roman*)]
  \item batch size ($n$) (1 -- 128), and
  \item the $l$, $m$, and $r$ dimensions (1 -- 1024).
\end{enumerate*}

\paragraph{Linear layers.}
For linear layers, we vary the
\begin{enumerate*}[label=(\roman*)]
  \item batch size (1~--~3500),
  \item input features (1~--~32768),
  \item output features (1~--~32768), and
  \item whether or not there is a bias weight.
\end{enumerate*}

\subsubsection{Data Preprocessing}
After collecting data on the GPUs, we build one dataset per operation by
\begin{enumerate*}[label=(\roman*)]
  \item adding the forward and backward execution times to arrive at a single
    execution time for each operation instance on a particular GPU, and
  \item attaching additional GPU hardware features to each of these data
    points.
\end{enumerate*}
We attach the GPU's
\begin{enumerate*}[label=(\roman*)]
  \item memory capacity and bandwidth;
  \item number of streaming multiprocessors (SMs); and
  \item peak FLOPS, as specified by the GPU manufacturer.
\end{enumerate*}

We present the characteristics of the final datasets in
Table~\ref{tbl:datasets}. We add four to the number of features to account for
the four GPU features (described above) that we add to each data point.
Similarly, in the dataset size column we show the total number of unique
operation configurations that we sample. We multiply by six because we make
measurements on six different GPUs.

\begin{table}[t]
\centering
\caption{A summary of the datasets used for our MLPs.}
\label{tbl:datasets}
\footnotesize
\begin{tabularx}{\columnwidth}{Xll}
  \toprule
  \textbf{Operation} & \textbf{Features} & \textbf{Dataset Size} \\
  \midrule
  2D Convolution & $7 + 4$ & $91,138 \times 6$ \\
  LSTM & $7 + 4$ & $124,176 \times 6$ \\
  Batched Matrix Multiply & $4 + 4$ & $131,022 \times 6$ \\
  Linear Layer & $4 + 4$ & $155,596 \times 6$ \\
  \bottomrule
\end{tabularx}
\end{table}

\subsubsection{Training}
We implement our MLPs using PyTorch. We train each MLP for \TrainEpochs{}
epochs using the Adam optimizer~\cite{adam-kingmaba15} with a learning rate of
\TrainLR{}, weight decay of \TrainWeightDecay{}, and a batch size of
\TrainBatchSize{} samples. We reduce the learning rate to \TrainLRSecond{}
after \TrainLRAdjustAfter{} epochs. We use the mean absolute percentage error
as our loss function:
\[
L = \frac{1}{n}
    \sum_{i=1}^{n}
    \left|
      \frac{\mathrm{predicted}_i - \mathrm{measured}_i}{\mathrm{measured}_i}
    \right|
\]
We assign 80\% of our data samples to the training set and the rest to our test
set. None of the configurations that we test on in Section~\ref{sec:evaluation}
appear in our training sets. We normalize the inputs by subtracting by the mean
and dividing by the standard deviation of the input features in our training
set.

\section{Evaluation}\label{sec:evaluation}
\thetool{} is meant to be used by deep learning researchers and practitioners
to predict the \emph{potential compute performance} of a given GPU so that they
can make \emph{informed} cost-efficient choices when selecting GPUs for
training.
Consequently, in our evaluation our goals are to determine
\begin{enumerate*}[label=(\roman*)]
  \item how \emph{accurately} \thetool{} can predict the training iteration
    execution time on GPUs with different architectures, and
  \item whether \thetool{} can correctly predict the relative cost-efficiency
    of different GPUs when used to train a given model.
\end{enumerate*}
Overall, we find that \thetool{} makes iteration execution time predictions
across pairs of \emph{six} different GPUs with an average error of
\EteAvgError{} on ResNet-50~\cite{resnet-he16}, Inception
v3~\cite{inception-szegedy15}, the Transformer~\cite{attention-vaswani17},
GNMT~\cite{gnmt-wu16}, and DCGAN~\cite{dcgan-radford16}.

\subsection{Methodology}\label{sec:methodology}
\begin{table}[t]
  \centering
  \caption{The GPUs we use in our evaluation.}\label{tbl:gpus}
  \footnotesize
  \setlength{\tabcolsep}{4pt} 
  \begin{tabularx}{\columnwidth}{Xlllll}
    \toprule
    \textbf{GPU} & \textbf{Generation} & \textbf{Mem.} & \textbf{Mem. Type} &
    \textbf{SMs} & \textbf{Rental Cost\footnotemark{}}\\
    \midrule
    P4000~\cite{p4000}   & \multirow{2}{*}{Pascal~\cite{pascal-nvidia}} & 8 GB & GDDR5~\cite{gddr5} & 14 & -- \\
    P100~\cite{p100}     & & 16 GB & HBM2~\cite{hbm2} & 56 & \$1.46/hr\\
    \midrule
    V100~\cite{v100}     & Volta~\cite{volta-nvidia} & 16 GB & HBM2 & 80 & \$2.48/hr \\
    \midrule
    2070~\cite{2070}     & \multirow{3}{*}{Turing~\cite{turing-nvidia}} & 8 GB & GDDR6~\cite{gddr6} & 36 & -- \\
    2080Ti~\cite{2080ti} & & 11 GB & GDDR6 & 68 & -- \\
    T4~\cite{t4}         & & 16 GB & GDDR6 & 40 & \$0.35/hr \\
    \bottomrule
  \end{tabularx}
\end{table}
\footnotetext{\href{https://cloud.google.com/compute/gpus-pricing}{Google Cloud
    pricing in us-central1, as of June 2021.}}

\begin{table}[t]
  \centering
  \caption{The machines we use in our evaluation.}\label{tbl:machines}
  \footnotesize
  \begin{tabularx}{\columnwidth}{Xllll}
    \toprule
    \textbf{CPU} & \textbf{Freq.} & \textbf{Cores} & \textbf{Main Mem.} &
    \textbf{GPU} \\
    \midrule
      Xeon E5-2680 v4~\cite{intel-e5-2680}  & 2.4 GHz & 14 & 128 GB & P4000 \\
      Ryzen TR 1950X~\cite{amd-1950x}       & 3.4 GHz & 16 & 16 GB  & 2070 \\
      EPYC 7371~\cite{amd-epyc-7371}        & 3.1 GHz & 16 & 128 GB & 2080Ti \\
    \bottomrule
  \end{tabularx}
\end{table}

\begin{table}[t]
  \centering
  \caption{The DNNs and training configurations we use.}\label{tbl:models}
  \footnotesize
  \begin{tabularx}{\columnwidth}{Xlll}
    \toprule
    {\bf Application} & {\bf Model} & {\bf Arch. Type} & {\bf Dataset} \\
    \midrule
    Image Classif. & ResNet-50~\cite{resnet-he16} & Convolution & ImageNet~\cite{imagenet} \\
    & Inception v3~\cite{inception-szegedy15} & & \\
    \midrule
    Machine Transl. & GNMT~\cite{gnmt-wu16} & Recurrent &
      \multirow{2}{*}{\shortstack{WMT'16~\cite{wmt16}\\\scriptsize (EN-DE)}} \\
    & Transformer~\cite{attention-vaswani17} & Attention & \\
    \midrule
    Image Gen. & DCGAN~\cite{dcgan-radford16} & Convolution & LSUN~\cite{lsun-yu15} \\
    \bottomrule
  \end{tabularx}
\end{table}

\paragraph{Hardware.}
In our experiments, we use the GPUs listed in Table~\ref{tbl:gpus}. For the
P4000, 2070, and 2080Ti we use machines whose configurations are listed in
Table~\ref{tbl:machines}. For the T4 and V100, we use \texttt{g4dn.xlarge} and
\texttt{p3.2xlarge} instances on AWS respectively~\cite{aws-instances}. For the
P100, we use Google Cloud's \texttt{n1-standard}
instances~\cite{gcp-n1standard} with 4 vCPUs and 15 GB of system memory.

\paragraph{Runtime environment.}
We run our experiments inside Docker containers~\cite{docker}. Our container
image uses Ubuntu 18.04~\cite{ubuntu}, CUDA 10.1~\cite{cuda}, and cuDNN
7~\cite{cudnn}. On cloud instances, we use the NVIDIA GPU Cloud Image, version
20.06.3~\cite{nvidia-ngc}. We use PyTorch 1.4.0~\cite{pytorch-paszke19} for all
experiments.

\paragraph{Models and datasets.}
We evaluate \thetool{} by predicting the training iteration execution time for
the models listed in Table~\ref{tbl:models} on different GPUs.
For ResNet-50 and Inception v3 we use stochastic gradient
descent~\cite{sgd-bottou10}.
We use Adam~\cite{adam-kingmaba15} for the rest of the models.
We use synthetic data (sampled from a normal distribution) of the \emph{same
  size} as samples from each dataset.\footnote{We verified that the training
  computation time does not depend on the \emph{values} of the data itself.}
For the machine translation models, we use a fixed sequence length of 50---the
longest sentence length typically used---to show how \thetool{} can make
predictions for a lower bound on the computational performance.

\paragraph{Metrics.}
In our experiments, we measure and predict the \emph{training iteration
  execution time}---the wall clock time it takes to perform one training step
on a batch of inputs. We use the training iteration execution time to compute
the training \emph{throughput} and \emph{cost-normalized throughput} for our
analysis. The training throughput is the batch size divided by the iteration
execution time. The cost-normalized throughput is the throughput divided by the
hourly cost of renting the hardware.

\paragraph{Measurements.}
We use CUDA events to measure the execution time of training iterations and DNN
operations. We run 3 warm up repetitions, which we discard, and then record the
average execution time over 3 further repetitions. We use CUPTI~\cite{cupti} to
measure a kernel's execution time.

\begin{figure*}[t]
  \begin{subfigure}[t]{0.5\textwidth}
    \includegraphics[width=\textwidth]{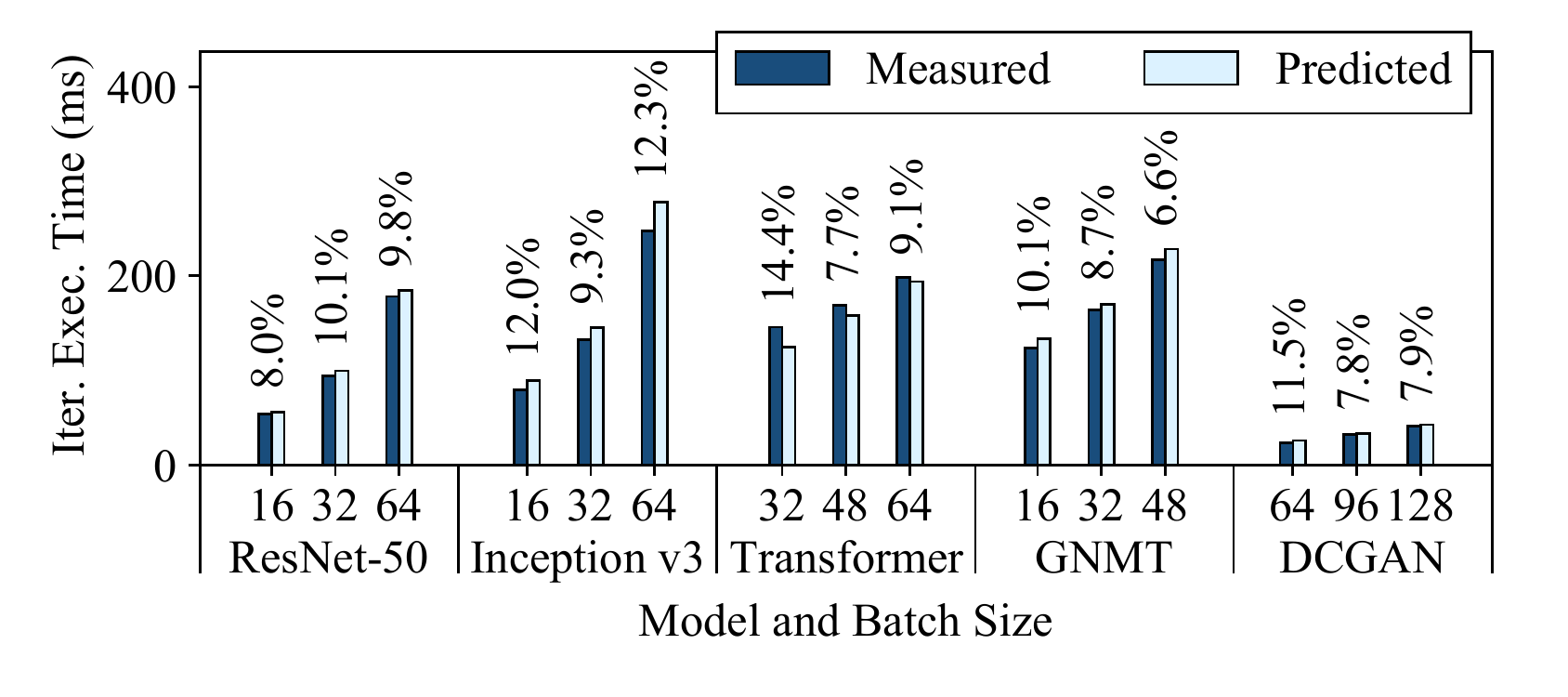}
    \vspace{-1.9em}
    \caption{Predictions onto the V100}
  \end{subfigure}
  \begin{subfigure}[t]{0.5\textwidth}
    \includegraphics[width=\textwidth]{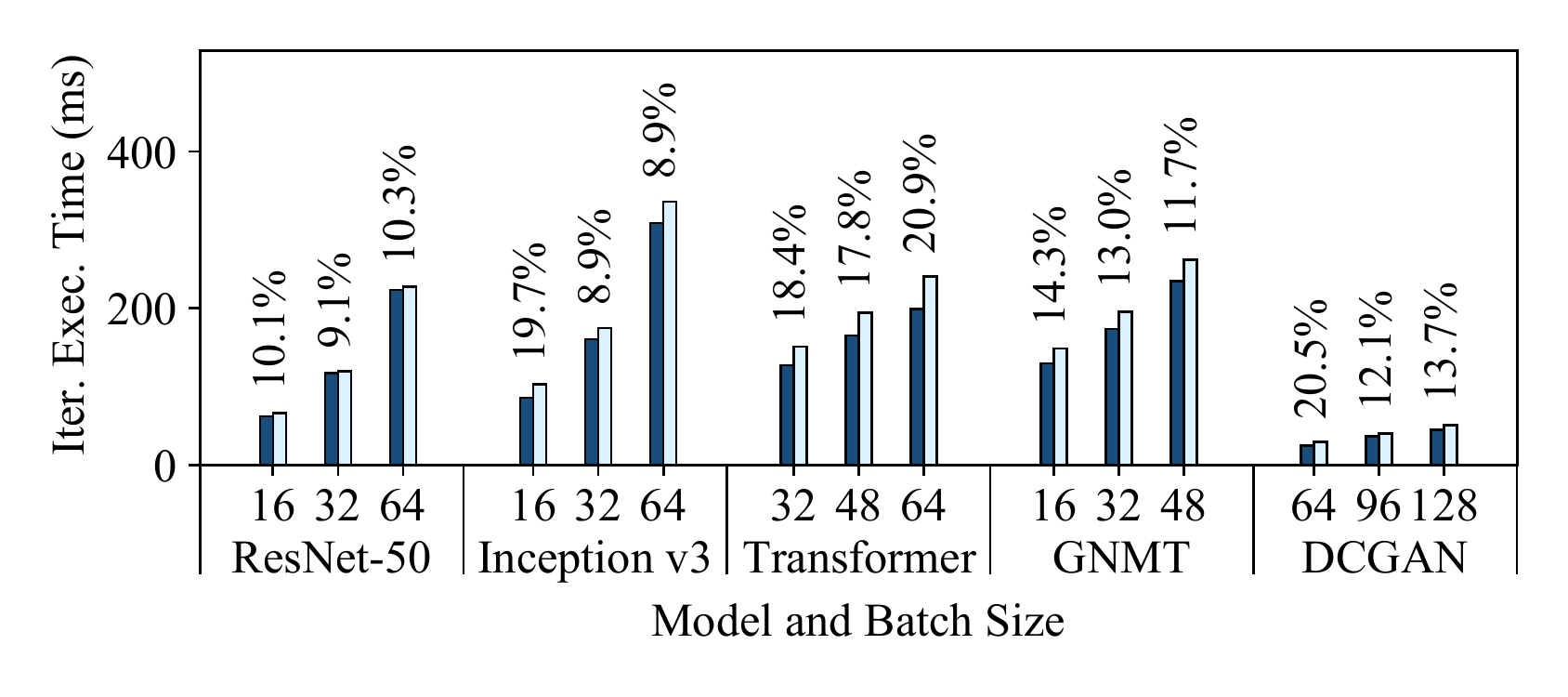}
    \vspace{-1.9em}
    \caption{Predictions onto the 2080Ti}
  \end{subfigure}

  \begin{subfigure}[t]{0.5\textwidth}
    \includegraphics[width=\textwidth]{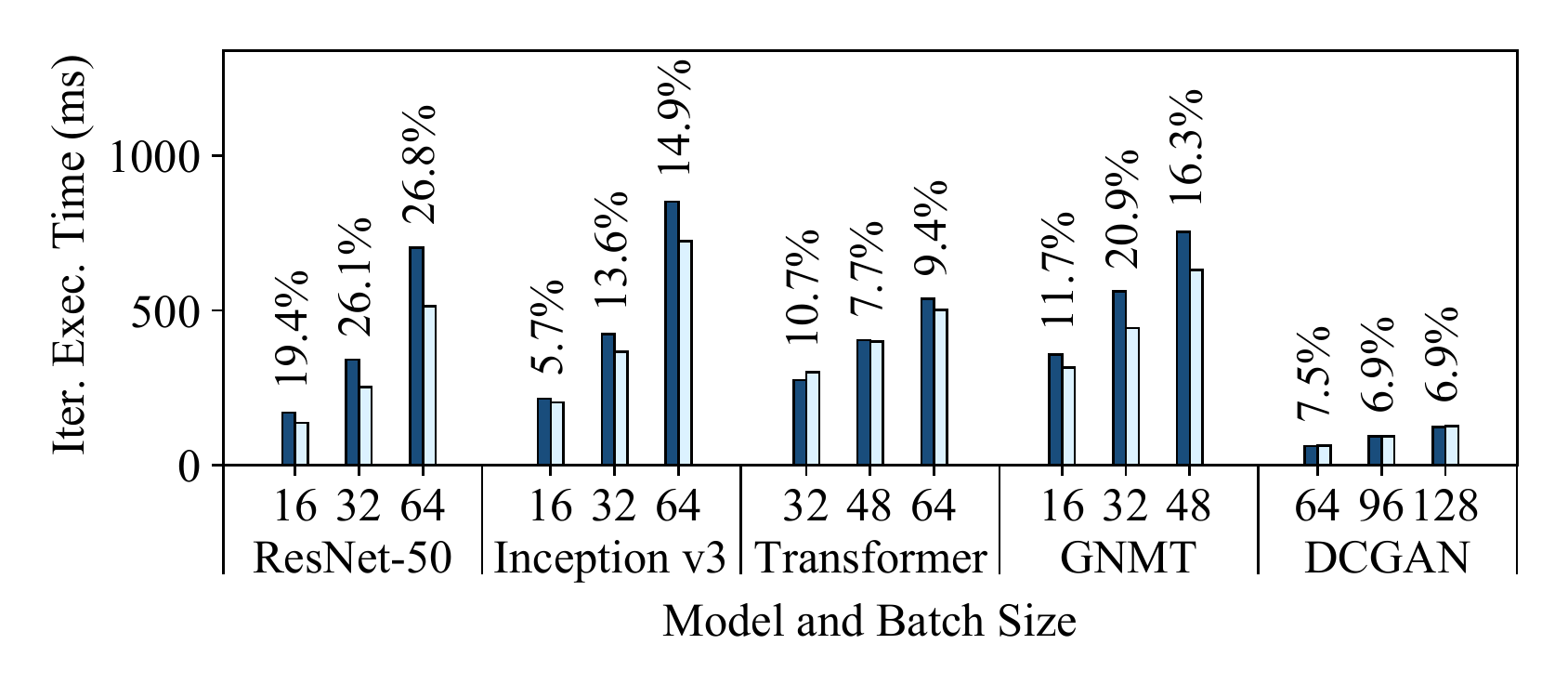}
    \vspace{-1.9em}
    \caption{Predictions onto the T4}
  \end{subfigure}
  \begin{subfigure}[t]{0.5\textwidth}
    \includegraphics[width=\textwidth]{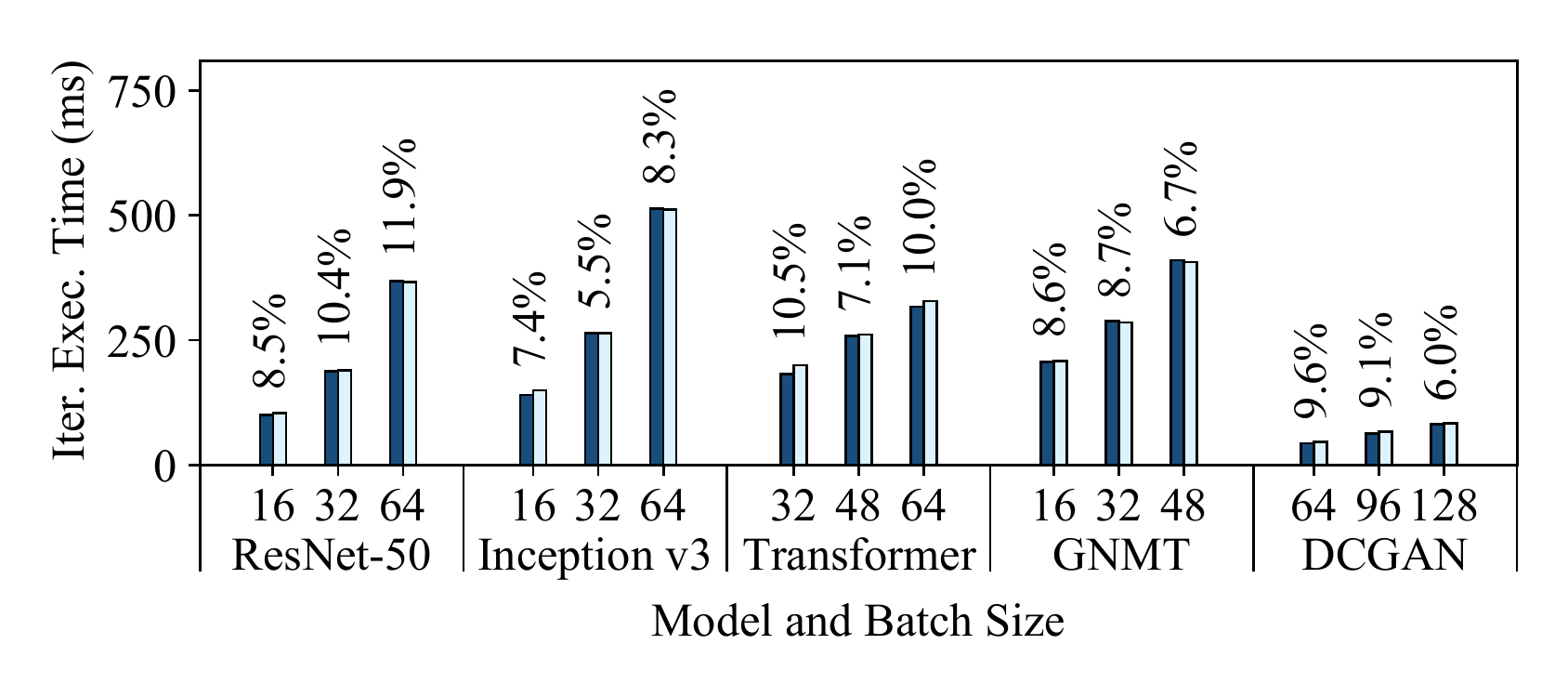}
    \vspace{-1.9em}
    \caption{Predictions onto the 2070}
  \end{subfigure}

  \begin{subfigure}[t]{0.5\textwidth}
    \includegraphics[width=\textwidth]{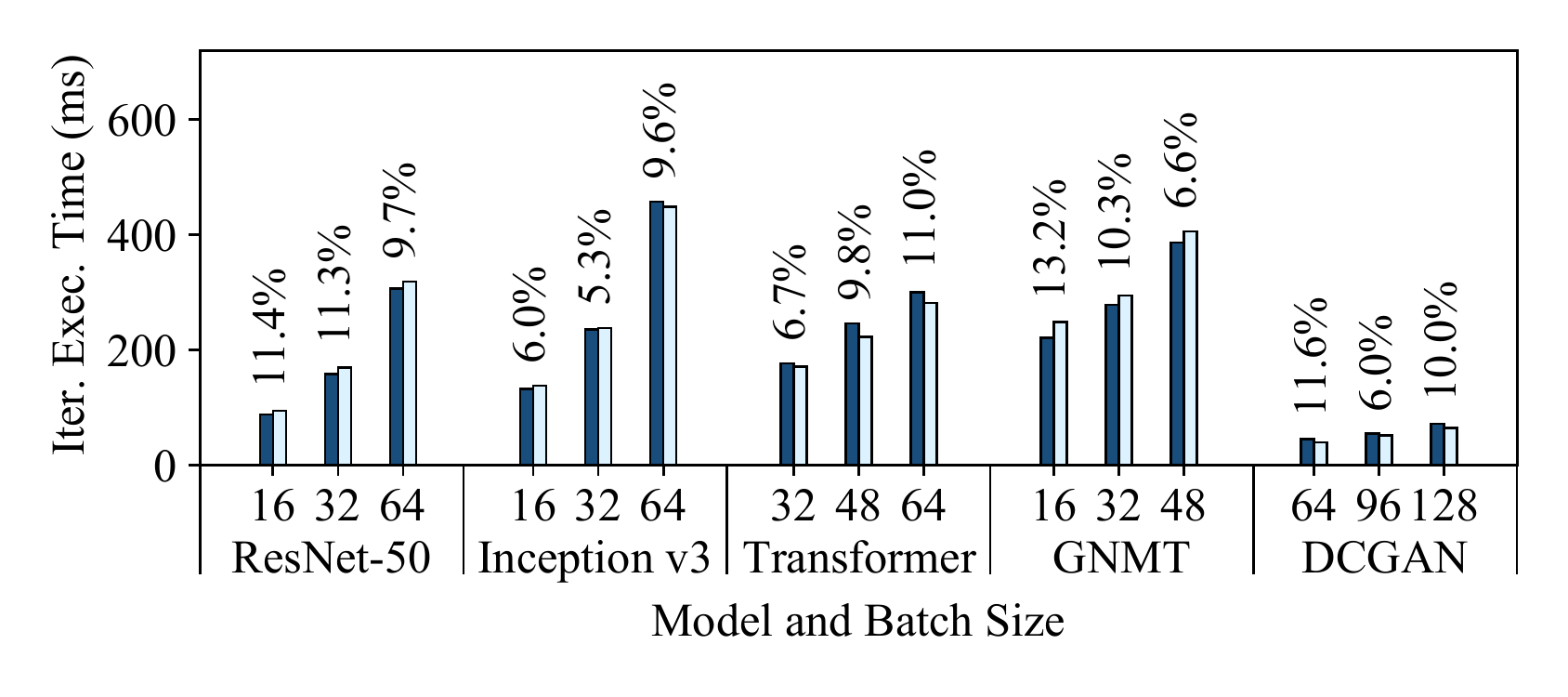}
    \vspace{-1.9em}
    \caption{Predictions onto the P100}
  \end{subfigure}
  \begin{subfigure}[t]{0.5\textwidth}
    \includegraphics[width=\textwidth]{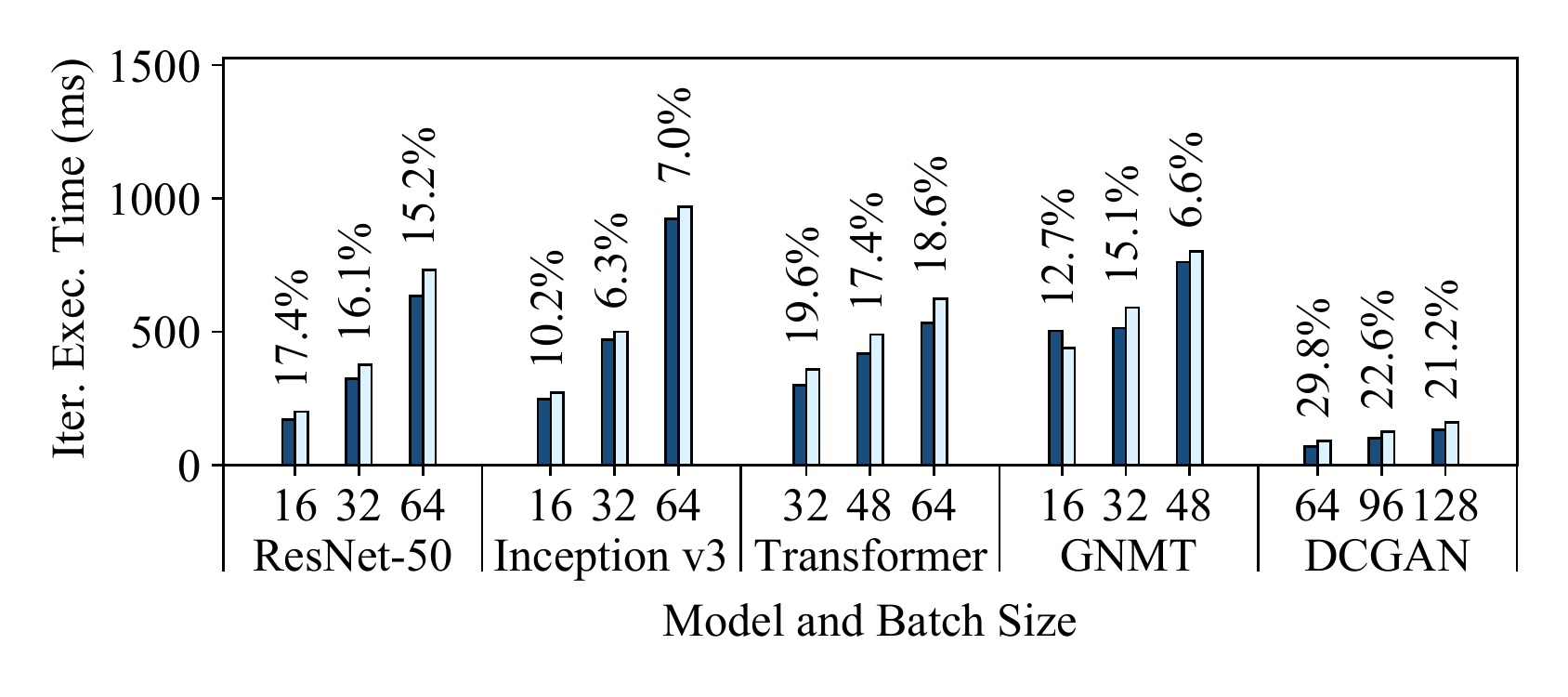}
    \vspace{-1.9em}
    \caption{Predictions onto the P4000}
  \end{subfigure}
  \caption{Iteration execution time predictions averaged across all other
    ``origin'' GPUs we evaluate.}
  \label{fig:e2e-main}
\end{figure*}

\begin{figure*}[t]
  \centering
  \includegraphics[width=\textwidth]{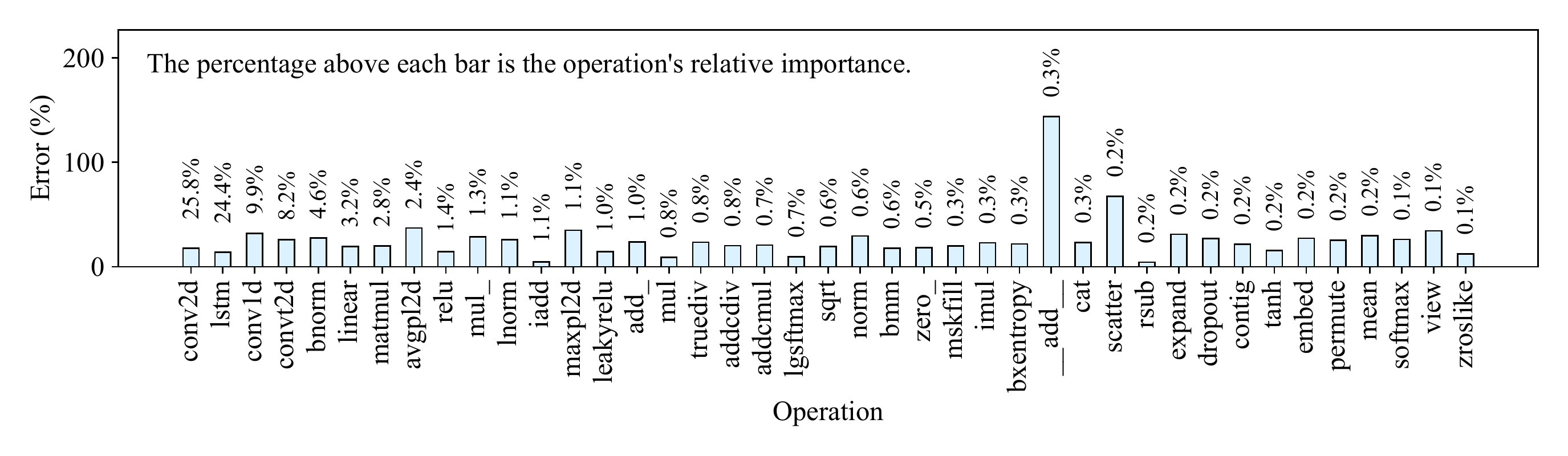}
  \vspace{-2.5em}
  \caption{Operation execution time prediction errors, with importance on top
    of each bar, averaged across all pairs of evaluated GPUs and models. The
    operation names have been shortened and we only show operations with an
    importance of at least 0.1\%.}
  \label{fig:op-breakdown}
  \vspace{-1em}
\end{figure*}

\subsection{How Accurate are \thetool{}'s
  Predictions?}\label{sec:evaluation-pred-acc}

To evaluate \thetool{}'s prediction accuracy, we use it to make training
iteration execution time predictions for ResNet-50, Inception v3, the
Transformer, GNMT, and DCGAN on all six GPUs listed in
Section~\ref{sec:methodology}. Recall that \thetool{} makes execution time
predictions by \emph{scaling} the execution time of a model and specific batch
size measured on one GPU (the ``origin'' GPU) to another (the ``destination''
GPU). As a result, we use all 30 possible (origin, destination) pairs of these
six GPUs in our evaluation.

\subsubsection{End-to-End Prediction Accuracy}
Figure~\ref{fig:e2e-main} shows \thetool{}'s prediction errors for these
aforementioned end-to-end predictions. Each subfigure shows the predictions for
all five models on a specific destination GPU. We make predictions for three
different batch sizes (shown on the figures) and plot both the predicted and
measured iteration execution times. Since we consider all possible pairs of our
six GPUs, for each destination GPU we plot the average predicted execution
times among the five origin GPUs. Similarly, we show the average prediction
error above each bar. From these figures, we can draw three major conclusions.

First, \thetool{} makes accurate end-to-end iteration execution time
predictions since the average prediction error across all GPUs and models is
\EteAvgError{}. The average prediction error across all ResNet-50, Inception
v3, Transformer, GNMT, and DCGAN configurations are \ResNetAvgError{},
\InceptionAvgError{}, \TransformerAvgError{}, \GNMTAvgError{}, and
\DCGANAvgError{} respectively.

Second, \thetool{} can predict the iteration execution time across GPU
\emph{generations}, which have different architectures, and across
\emph{classes} of GPUs. The GPUs we use span three generations
(Pascal~\cite{pascal-nvidia}, Volta~\cite{volta-nvidia}, and
Turing~\cite{turing-nvidia}) and include desktop, professional workstation, and
server-class GPUs.

Third, \thetool{} is \emph{general} since it supports different types of DNN
architectures. \thetool{} works with convolutional neural networks (e.g.,
ResNet-50, Inception v3, DCGAN), recurrent neural networks (e.g., GNMT), and
other neural network architectures such as the attention-based Transformer. In
particular, \thetool{} makes accurate predictions for ResNet, Inception, and
DCGAN despite the significant differences in their architectures; ResNet has a
``straight-line'' computational graph, Inception has a large ``fanout'' in its
graph, and DCGAN is a generative-adversarial model.

\subsubsection{Prediction Error Breakdown}
Figure~\ref{fig:op-breakdown} shows a breakdown of the prediction errors for
the execution time of individual operations, which are listed on the $x$-axis.
The operations predicted using the MLP predictors are shown on the left
(\texttt{conv2d}, \texttt{lstm}, \texttt{bmm}, and \texttt{linear}).
\Kernelsim{} is used to predict the rest of the operations. Above each bar, we
also show the \emph{importance} of each operation as a percentage of the
iteration execution time, averaged across all five DNNs. The prediction errors
are averaged among all pairs of the six GPUs that we evaluate and among
ResNet-50, Inception v3, the Transformer, GNMT, and DCGAN. From this figure, we
can draw two major conclusions.

First, MLP predictors can be used to make accurate predictions for
kernel-varying operations as the average error among the \texttt{conv2d},
\texttt{lstm}, \texttt{bmm}, and \texttt{linear} operations is
\MLPOpAvgError{}. Second, \kernelsim{} can make accurate predictions for
important operations; the average error for \kernelsim{} predictions is
\WaveOpAvgError{}. Although \kernelsim{}'s predictions for some operations
(e.g., \texttt{\_\_add\_\_}, \texttt{scatter}) have high errors, these
operations do not make up a significant proportion of the training iteration
execution time (having an overall importance of at most 0.3\%).

\subsubsection{Prediction Contribution Breakdown}
We also examine how \kernelsim{} and the MLPs each contribute to making
\thetool{}'s end-to-end predictions. In our evaluation, \thetool{} uses
\kernelsim{} for \WaveOperationCountProportion{} of the unique operations; it
uses MLPs for the other \MLPOperationCountProportion{}. In contrast, when
looking at execution time, \thetool{} uses \kernelsim{} to predict
\WaveTimeProportion{} of an iteration's execution time on average; it uses MLPs
for the other \MLPTimeProportion{}.

These breakdowns show that \emph{both} \kernelsim{} and the MLPs contribute
non-trivially to \thetool{}'s predictions---each is responsible for roughly
half of an iteration's execution time. Additionally, the unique operation
breakdown shows that most operations are predicted using \kernelsim{}. This
observation highlights a strength of \thetool{}'s hybrid approach of using
both \kernelsim{} and MLPs: most operations can be automatically predicted
using \kernelsim{}; MLPs only need to be trained for a few kernel-varying
operations.

\subsubsection{MLPs: How Many Layers?}
\begin{figure}
  \includegraphics[width=\columnwidth]{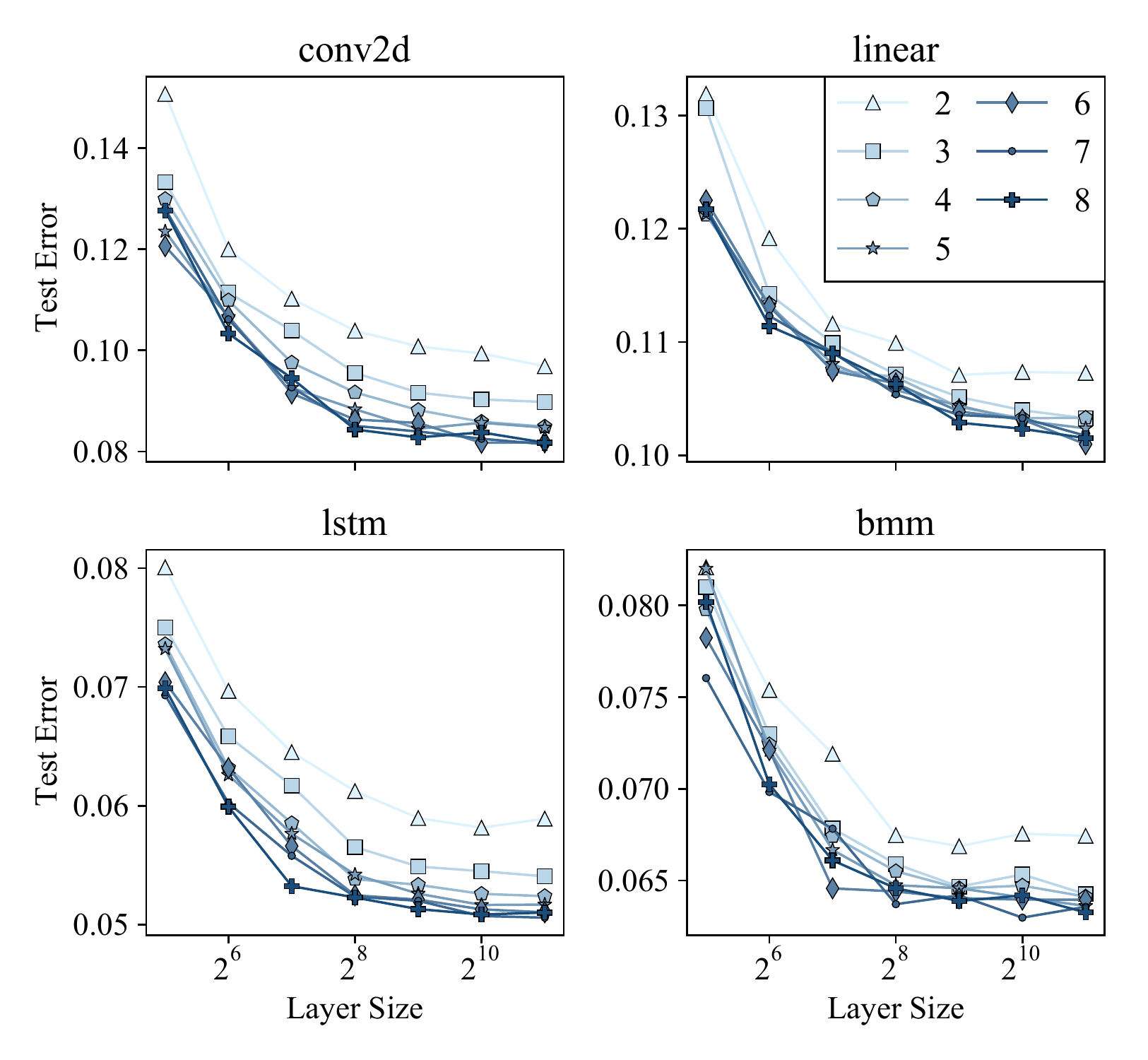}
  \vspace{-2.5em}
  \caption{Test error as we vary the number of layers and their sizes in each
    MLP. The $x$-axis is in a logarithmic scale.}
  \label{fig:ablation}
\end{figure}

In all our MLPs, we use eight hidden layers, each of size 1024. To better
understand how the number of layers affects the MLPs' prediction accuracy, we
also conduct a sensitivity study where we vary the number of hidden layers in
each MLP (2 to 8) along with their size (powers of two: $2^5$ to $2^{11}$).
Figure~\ref{fig:ablation} shows each MLP's test mean absolute percentage error
after being trained for 80 epochs. From this figure we can draw two major
conclusions.

First, increasing the number of layers and their sizes leads to lower test
errors. Increasing the size of each layer beyond $2^9$ seems to lead to
diminishing returns on each operation. Second, the MLPs for all four operations
appear to follow a similar test error trend. Based on these results, we can
also conclude that using eight hidden layers is a reasonable choice.

\subsection{Does \thetool{} Lead to Correct Decisions?}\label{sec:case-studies}
One of \thetool{}'s primary use cases is to help deep learning users make
\emph{informed} and \emph{cost-efficient} GPU selections. In the following two
case studies, we demonstrate how \thetool{} can make cost-efficiency
predictions that empower users to make correct selections according to their
needs.

\subsubsection{Case Study 1: Should I Rent a Cloud GPU?}
As mentioned in Section~\ref{sec:introduction}, one scenario a deep learning
user may face is deciding whether to rent GPUs in the cloud for training or to
stick with a GPU they already have locally (e.g., in their desktop). For
example, suppose a user has a P4000 in their workstation and they want to
decide whether to rent a P100, T4, or V100 in the cloud to train GNMT.

\begin{figure}
  \begin{subfigure}[t]{\columnwidth}
    \includegraphics[width=\textwidth]{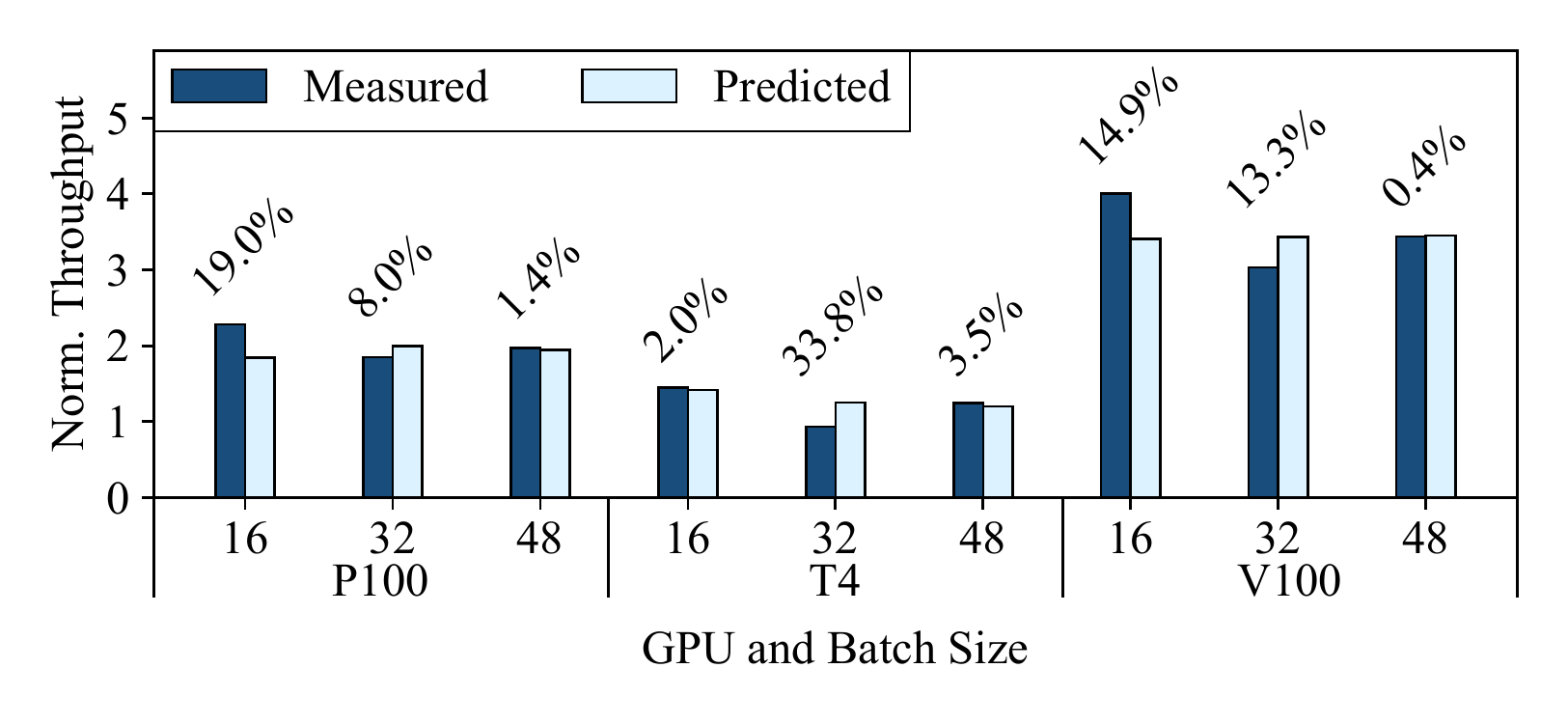}
    \vspace{-1.5em}
    \caption{GNMT training throughput normalized to the P4000}
    \label{fig:case-study-norm-p4000}
  \end{subfigure}
  \begin{subfigure}[t]{\columnwidth}
    \includegraphics[width=\textwidth]{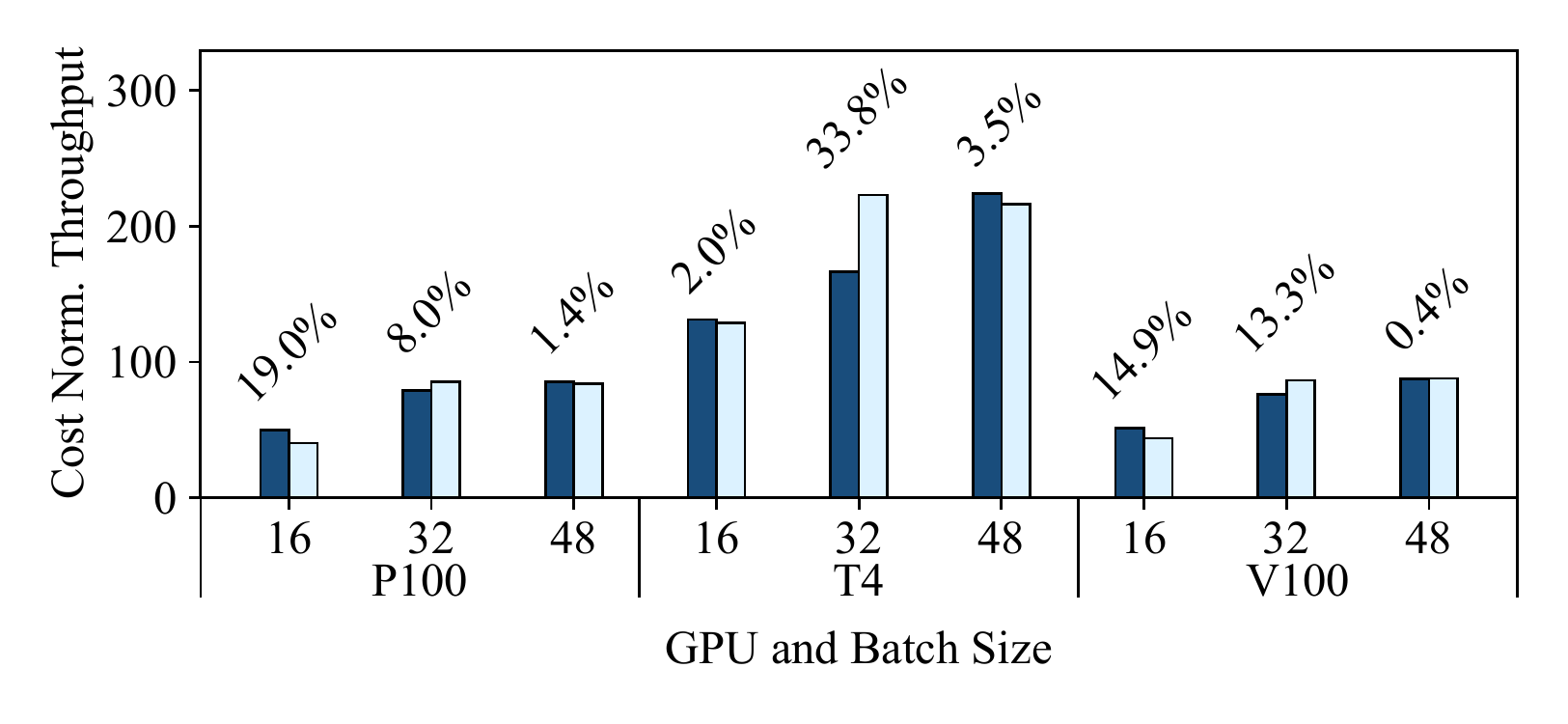}
    \vspace{-1.5em}
    \caption{GNMT cost normalized throughput}
    \label{fig:case-study-norm-cost}
  \end{subfigure}
  \caption{\thetool{}'s GNMT training throughput predictions for cloud GPUs,
    made using a P4000. The percentage error is shown above each prediction.}
  \label{fig:case-study}
\end{figure}

With \thetool{}, they can use their P4000 to make \emph{predictions} about
the computational performance of each cloud GPU to help them make this decision
in an informed way. Figure~\ref{fig:case-study-norm-p4000} shows \thetool{}'s
throughput predictions for GNMT for the P100, T4, and V100 normalized to the
training throughput on the P4000. Additionally,
Figure~\ref{fig:case-study-norm-cost} shows \thetool{}'s predicted training
throughputs normalized by each cloud GPU's rental costs on Google Cloud as
shown in Table~\ref{tbl:gpus}. Note that
\begin{enumerate*}[label=(\roman*)]
  \item we make all these predictions with the P4000 as the origin device,
  \item we make our ground truth measurements on Google Cloud instances, and
  \item one can also use \thetool{} for a similar analysis for other cloud
    providers.
\end{enumerate*}
From these results, the user can make two observations.

First, both the P100 and V100 offer training throughput speedups over the P4000
(up to $2.3\times$ and $4.0\times$ respectively) whereas the T4 offers marginal
throughput speedups (up to $1.4\times$). However, second, the user would also
discover that the T4 is more \emph{cost-efficient} to rent when compared to the
P100 and V100 as it has a higher cost-normalized throughput. Therefore, if the
user wanted to optimize for maximum computational performance, they would
likely choose the V100. But if they were not critically constrained by time and
wanted to optimize for cost, sticking with the P4000 or renting a T4 would be a
better choice.

\thetool{} makes these predictions accurately, with an average error of
\CloudCaseAvgError{}. We also note that despite any prediction errors,
\thetool{} still \emph{correctly} predicts the relative ordering of these three
GPUs in terms of their throughput and cost-normalized throughput. For example,
in Figure~\ref{fig:case-study-norm-cost}, \thetool{} correctly predicts that the
T4 offers the best cost-normalized throughput on all three batch sizes. These
predictions therefore allow users to make correct decisions based on their
needs (optimizing for cost or pure performance).

\subsubsection{Case Study 2: Is the V100 Always Better?}

In the previous case study, \thetool{} correctly predicts that the V100
provides the best performance despite not being the most cost-efficient to
rent. This conclusion may lead a na\"ive user to believe that the V100 always
provides better training throughput over other GPUs, given that it is the most
advanced and expensive GPU available in the cloud to rent.\footnote{This is
  true except for the new A100s, which have only recently become publicly
  available in the cloud.} In this case study, we show how \thetool{} can help
a user recognize when the V100 does not offer significant performance benefits
for their model.

\begin{figure}
  \hspace{-1em}
  \includegraphics[width=1.05\columnwidth]{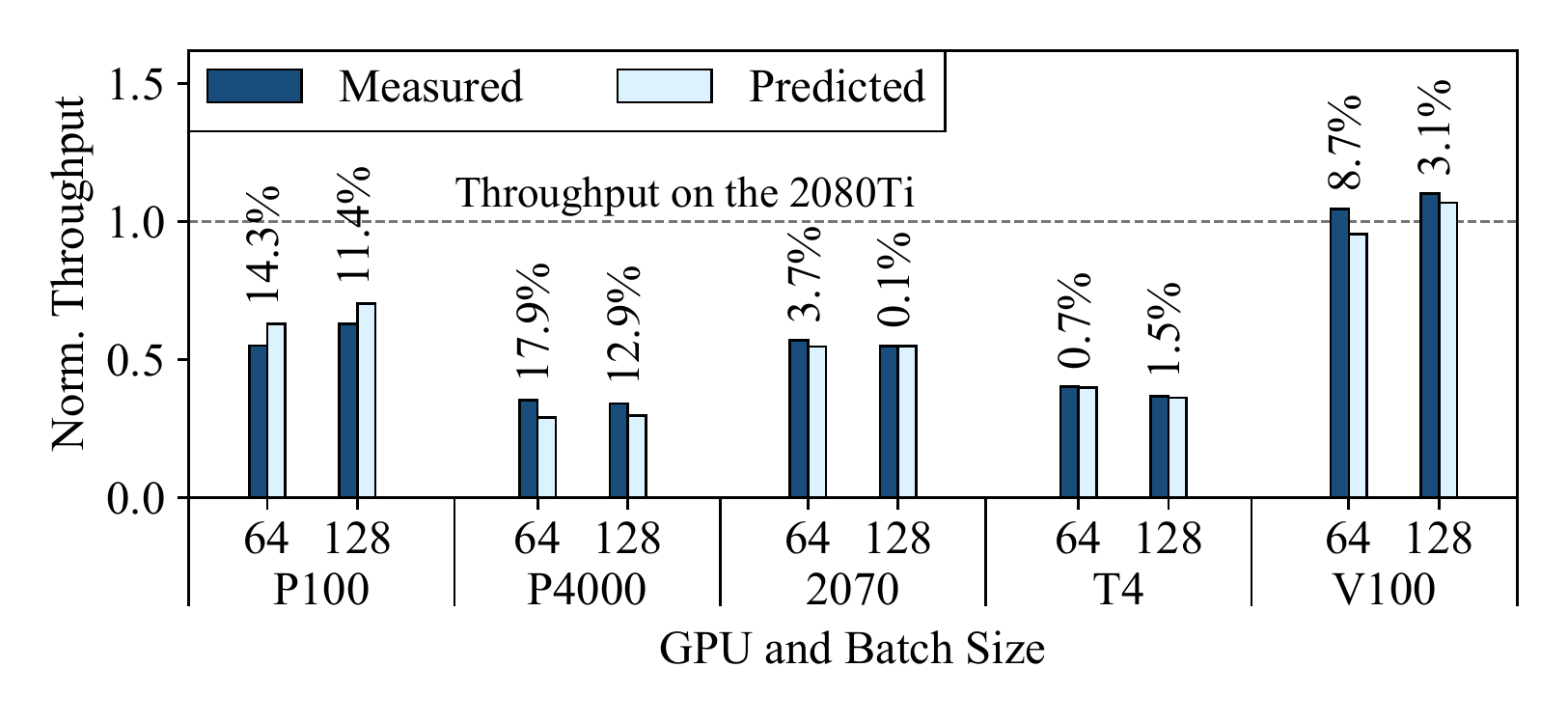}
  \vspace{-2.2em}
  \caption{Predicted and measured DCGAN training throughput normalized to the
    2080Ti, with prediction errors above each bar. \thetool{} correctly
    predicts that the V100's performance is not significantly better than the
    2080Ti.}
  \label{fig:dcgan-case-study}
\end{figure}

Suppose a user wants to train DCGAN and already has a 2080Ti that they can use.
They want to find out if they should use a different GPU to get better
computational performance (training throughput). They can use \thetool{} to
predict the training throughput on other GPUs.
Figure~\ref{fig:dcgan-case-study} shows \thetool{}'s throughput predictions
along with the measured throughput, normalized to the 2080Ti's training
throughput. Note that we use a batch size of 64 as it is the default batch size
in the DCGAN reference implementation~\cite{dcgan-code} and 128 because it is
the size reported by the authors in their paper~\cite{dcgan-radford16}.

From this figure, the user would conclude that they should stick to using their
2080Ti as the V100 would not be worth renting. The V100 offers marginal
throughput improvements over the 2080Ti (\GeneralCaseMaxSpeedup{}) while the
P100, P4000, 2070, and T4 all do not offer throughput improvements at all. The
reason the V100 does not offer any significant benefits over the 2080Ti despite
having more computational resources (Table~\ref{tbl:gpus}) is that DCGAN is a
``computationally lighter'' model compared to GNMT and so it does not really
benefit from a more powerful GPU. \thetool{} makes these predictions
accurately, with an average error of \GeneralCaseAvgError{}.

\paragraph{Summary.}
These case studies show examples of situations where
\begin{enumerate*}[label=(\roman*)]
  \item the GPU offering the highest training throughput is \emph{not} the same
    as the most cost-efficient GPU, and where
  \item the V100 does \emph{not} offer significantly better performance when
    compared to a desktop-class GPU (the 2080Ti).
\end{enumerate*}
Notably, in both case studies, \thetool{} \emph{correctly} predicts each of
these findings.
As a result, deep learning researchers and practitioners can rely on \thetool{}
to help them make correct cost-efficient GPU selections according to their
needs.

\section{Discussion}
In this section, we discuss how \thetool{} can be extended to support
additional
\begin{enumerate*}[label=(\roman*)]
  \item training setups, and
  \item deep learning frameworks.
\end{enumerate*}
In doing so, we also highlight opportunities for future work and describe some
challenges and opportunities associated with supporting additional hardware
accelerators

\subsection{Additional Training Setups}
\thetool{} is designed to make accurate cross-GPU execution time predictions
for DNN training. However, users may also face situations where they need
performance predictions for more complex training setups such as
\begin{enumerate*}[label=(\roman*)]
  \item distributed training~\cite{ddnns-dean12},
  \item mixed precision training~\cite{mixedprecision-micikevicius18}, or
  \item needing predictions for batch sizes larger than what can fit on the
    origin GPU.
\end{enumerate*}
In this subsection, we outline how \thetool{} could be extended to support
these setups.

\subsubsection{Distributed Training}
Predicting the execution time of a distributed training iteration generally
reduces to predicting
\begin{enumerate*}[label=(\roman*)]
  \item the computation time on the cluster's GPUs,
  \item the communication time among the GPUs and/or nodes, and
  \item how the communication overlaps with the computation.
\end{enumerate*}
For data parallel training~\cite{ddnns-dean12}, several prior works present
techniques for predicting the data parallel iteration execution time given the
execution time on a single
GPU~\cite{daydream-zhu20,cnniterpred-pei19,paleo-qi17} (i.e., tasks (ii) and
(iii)). \thetool{}'s computation predictions (task (i)) could be used as an
input to these existing techniques.

For more complex distribution schemes such as model
parallel~\cite{ddnns-dean12} and pipeline parallel
training~\cite{genpipeline-narayanan19,gpipe-huang19}, \thetool{} could still
be used for task (i), but the user would need to split up their model based on
the distributed partitioning scheme before profiling it with \thetool{}.
However, for tasks (ii) and (iii), new prediction techniques would need to be
developed. This is something we leave to future work.

\subsubsection{Mixed Precision Training}
The Daydream paper~\cite{daydream-zhu20} presents a technique for predicting
the performance benefits of switching from single to mixed precision training
on the \emph{same} fixed GPU.
If users want to know about the performance benefits of mixed precision
training on a \emph{different} GPU, they can use Daydream's technique in
conjunction with \thetool{}.

To show that this combined approach can work in practice, we use a P4000 to
predict the execution time of a ResNet-50 mixed precision training iteration on
the 2070 and 2080Ti.\footnote{We use the same experimental setup and batch
  sizes as described and shown in Section~\ref{sec:methodology} and
  Figure~\ref{fig:e2e-main}. We compare our iteration execution time
  predictions against training iterations performed using PyTorch's automatic
  mixed precision module.}
On the P4000, we first use \thetool{} to predict the single precision iteration
execution time on the 2070 and 2080Ti.
Then, we apply Daydream's technique to translate these predicted single
precision execution times into mixed precision execution times.
We also repeat this experiment between the 2070 and 2080Ti.
Overall, we find that this approach has an average error of
\MixedPrecisionAvgError{} for predictions onto the 2070 and 2080Ti.

To distinguish between the errors introduced by \thetool{} versus Daydream, we
also apply Daydream's technique to the measured (i.e., ground truth) single
precision iteration execution times.
We find that Daydream's technique alone has an average error of
\MixedPrecisionDaydreamOnlyAvgError{} for the 2070 and 2080Ti.
Thus we believe the additional error introduced by also using \thetool{}
is reasonable, given the extra functionality.
So overall, we conclude that \thetool{} with Daydream should be able to
effectively support mixed precision predictions on other GPUs.

\subsubsection{Larger Batch Sizes}
Recall that \thetool{}'s iteration execution time predictions are for a model
and a specific batch size. This means that the origin GPU must be able to run a
training iteration with the desired batch size (for \thetool{}'s profiling
pass).

One potential approach to making predictions for batch sizes larger than what
can run on the origin GPU is as follows. First, use \thetool{} to make
iteration execution time predictions for multiple (e.g., three) different batch
sizes that \emph{do} fit on the origin GPU. Then, build a linear regression
model on these predicted values to \emph{extrapolate} to larger batch sizes.
This approach is based on our prior work, where we observed an often linear
relationship between the iteration execution time and batch
size~\cite{skyline-yu20}. We leave the handling of models where only one batch
size fits on the origin GPU to future work.
\vspace{-0.3em}

\subsection{Additional Deep Learning Frameworks}
Recall that \thetool{} predicts the execution time of operations using either
\begin{enumerate*}[label=(\roman*)]
  \item \kernelsim{} or
  \item pre-trained MLPs,
\end{enumerate*}
depending on whether the operation is kernel-alike or kernel-varying.
Therefore, as long as \thetool{} has information about a DNN's operations and
their parameters (e.g., batch size, number of channels), \thetool{} will be
able to apply its techniques to make execution time predictions for a different
GPU.
Ultimately this means that adding support for other deep learning frameworks
(e.g., TensorFlow or MXNet) boils down to extracting the underlying operations
that run during a training iteration and sending the operations to \thetool{}
(i.e., extracting the computation graph).
Since the other major deep learning frameworks (TensorFlow and MXNet) both
already use computation graphs
internally~\cite{tensorflow-abadi16,mxnet-chen15}, we believe that adding
support for them would be straightforward to implement.
\vspace{-0.3em}

\subsection{Additional Hardware Accelerators}
As described in Section~\ref{sec:introduction}, there are also other hardware
options available beyond GPUs that can be used for training (e.g., the
TPU~\cite{tpu-jouppi17}, AWS Trainium~\cite{trainium}, and
Gaudi~\cite{habana}).
Therefore, a natural opportunity for future work is to explore execution time
predictions for these other hardware accelerators.
We outline two challenges that arise when going beyond GPUs, as well as two
examples of ways that \thetool{}'s guiding principles can be applied to these
prediction tasks.

\paragraph{Challenges.}
First, specialized deep learning accelerators may have a different hardware
architecture when compared to GPUs---necessitating different performance
modeling techniques.
For example, the TPU uses a systolic array~\cite{tpu-jouppi17,systolic-brent84}
whereas GPUs are general-purpose SIMT processors~\cite{cudaprog}.
Second, accelerators such as the TPU rely on \emph{tensor compilers} (e.g.,
XLA~\cite{xla} or JAX~\cite{jax}) to produce executable code from the
high-level DNN model code written by an end-user.
The compiler may apply optimizations that change the operations.
These changes make the high-level operation-based analysis that \thetool{}
performs more difficult to realize.

\paragraph{Opportunities.}
Despite these challenges, we believe that there are also opportunities to apply
\thetool{}'s key idea of leveraging runtime-based information from one
accelerator to predict the execution time on a different accelerator.
For example, as of June 2021, Google makes two versions of the TPU available
for rent (v2 and v3)~\cite{tpu-versions} and has announced the
v4~\cite{tpuv4-blog}.
Execution times measured on the TPU v2 could potentially be used to make
execution time predictions on the v3 and v4 and vice-versa.
Similarly, assuming that the AWS Trainium also uses a systolic
array,\footnote{The AWS Inferentia~\cite{inferentia,inferentia-zheng20}, a
  related accelerator, uses a systolic array
  architecture~\cite{inferentia-announce}. So we believe that this is a
  reasonable assumption to make.} it may also be possible to leverage execution
time measurements on the TPU to make execution time predictions for the
Trainium and vice-versa.
\vspace{-0.2em}

\section{Related Work}
\vspace{-0.15em}
The key difference between \thetool{} and existing DNN performance modeling
techniques for GPUs~\cite{paleo-qi17,predictdnncost-justus18,cnniterpred-pei19}
is in how \thetool{} makes execution time predictions.
\thetool{} takes a hybrid \emph{runtime-based approach}; it uses information
recorded at runtime on one GPU along with hardware characteristics to
\emph{scale} the measured kernel execution times onto different
GPUs through either
\begin{enumerate*}[label=(\roman*)]
  \item \kernelsim{}, or
  \item pre-trained MLPs.
\end{enumerate*}
In contrast, existing techniques use analytical
models~\cite{paleo-qi17,cnniterpred-pei19} or rely \emph{entirely} on machine
learning techniques~\cite{predictdnncost-justus18}.
The key advantage of \thetool{}'s hybrid scaling approach is that \kernelsim{}
works ``out of the box'' for all kernel-alike operations (i.e., operations
implemented using the same kernels on different GPUs).
Ultimately, this advantage means that new analytical or machine learning models
do not have to be developed each time a new kernel-alike operation is
introduced.

\paragraph{DNN performance models for different hardware.}
There exists prior work on performance models for DNN training on
GPUs~\cite{paleo-qi17,predictdnncost-justus18,cnniterpred-pei19},
CPUs~\cite{inteldnn-viebke2019}, and TPUs~\cite{learnedtpu-kaufman21}.
As described above, \thetool{} is fundamentally different from these works
because it takes a hybrid \emph{runtime-based approach} when making execution
time predictions.
For example, Paleo~\cite{paleo-qi17}
\begin{enumerate*}[label=(\roman*)]
  \item makes DNN operation execution time predictions using \emph{analytical
      models} based on the number of floating point operations (FLOPs) in a DNN
    operation, and
  \item uses heuristics to select the kernels used to implement kernel-varying
    operations (e.g., convolution).
\end{enumerate*}
However, an operation's execution time is not solely determined by its number
of FLOPs, and using heuristics to select an analytical model cannot always
capture kernel-varying operations correctly.
This is because proprietary closed-source kernel libraries (e.g.,
cuDNN~\cite{cudnn,cudnn-chetlur14}, cuBLAS~\cite{cublas}) may select different
kernel(s) to use by running benchmarks on the target
GPU~\cite{cudnnperf-jorda19,cudnnfind}.

\paragraph{Performance models for compilers.}
A complementary body of work on performance modeling is motivated by the needs
of compilers: predicting how \emph{different implementations} of some
high-level functionality perform on the \emph{same hardware}.
These models were developed to aid in compiling high-performance
\begin{enumerate*}[label=(\roman*)]
  \item graphics pipelines~\cite{halidemodel-adams19},
  \item CPU code~\cite{ithemal-mendis19}, and
  \item tensor operators for deep learning
    accelerators~\cite{tvmmodel-chen18,learnedtpu-kaufman21}.
\end{enumerate*}
These models have fundamentally different goals compared to \thetool{}, which
is a technique that predicts the performance of \emph{different GPUs} running
the \emph{same high-level code}.

\paragraph{General scalability predictions.}
\Kernelsim{} is similar in spirit to ESTIMA~\cite{estima-chatzopoulos16}, since
both use the idea of making measurements on one system to make performance
predictions for a different system. However, ESTIMA is a scalability predictor
for CPU programs. \Kernelsim{} instead targets GPU kernels, which run under a
different execution model when compared to CPU programs.

\paragraph{Repetitiveness of DNN training.}
Prior work leverages the repetitiveness of DNN training computation to optimize
distributed
training~\cite{flexflow-jia19,parallax-kim19,genpipeline-narayanan19}, schedule
jobs in a
cluster~\cite{gandiva-xiao18,heterodlsched-narayanan20,gandivafair-chaudhary20},
and to apply DNN compiler optimizations~\cite{astra-sivathanu19}.
The key difference between these works and \thetool{} is that they apply
optimizations on the \emph{same} hardware configuration. \thetool{} exploits
the repetitiveness of DNN training to make performance predictions on
\emph{different} hardware configurations.

\paragraph{DNN benchmarking.}
A body of prior work focuses on benchmarking DNN
training~\cite{fathom-adolf16,tbd-zhu18,dawn-coleman17,mlperf}.
While these works provide DNN training performance insights, they do so only
for a \emph{fixed} set of DNNs and hardware configurations.
In contrast, \thetool{} analyzes DNNs in \emph{general} and provides
performance \emph{predictions} on different GPUs to help users make informed
GPU selections.
\vspace{-0.3em}

\section{Conclusion}
We present \emph{\thetool{}}: a new runtime-based library that uses
\kernelsim{} and MLPs as execution time predictors to help deep learning
researchers and practitioners make \emph{informed cost-efficient} decisions
when selecting a GPU for DNN training.
The key idea behind \thetool{} is to leverage information collected at runtime
on one GPU to help predict the execution time of a DNN training iteration on a
different GPU.
We evaluate \thetool{} and find that it makes cross-GPU iteration execution
time predictions with an overall average error of \EteAvgError{} on ResNet-50,
Inception v3, the Transformer, GNMT, and DCGAN.
Finally, we present two case studies where \thetool{} correctly predicts that
\begin{enumerate*}[label=(\roman*)]
  \item optimizing for cost-efficiency would lead to selecting a different GPU
    for GNMT, and
  \item that the V100 does not offer significant performance benefits over a
    common desktop-class GPU (the 2080Ti) for DCGAN.
\end{enumerate*}
We have also open sourced \thetool{} (\HabitatCodeLink{}) to benefit both the
deep learning and systems
communities~\cite{habitat-code-yu21,habitat-models-yu21}.

\section*{Acknowledgments}
We thank our shepherd, Marco Canini, and the anonymous reviewers for their
feedback.
We also thank (in alphabetical order)
Moshe Gabel,
James Gleeson,
Anand Jayarajan,
Xiaodan Tan,
Alexandra Tsvetkova,
Shang Wang,
Qiongsi Wu, and
Hongyu Zhu.
We thank all members of the
\href{https://www.cs.toronto.edu/ecosystem/}{EcoSystem research group} for the
stimulating research environment they provide.
This work was supported by a QEII-GSST, Vector Scholarship in Artificial
Intelligence, Snap Research Scholarship, and an NSERC Canada Graduate
Scholarship -- Master's (CGS M). This work was also supported in part by the
NSERC Discovery grant, the Canada Foundation for Innovation JELF grant, the
Connaught Fund, an Amazon Research Award, and a Facebook Faculty Award.
Computing resources used in this work were provided, in part, by the Province
of Ontario, the Government of Canada through CIFAR, and companies sponsoring
the Vector Institute
\href{https://www.vectorinstitute.ai/partners}{www.vectorinstitute.ai/partners}.

\bibliography{references/deep-learning,references/ml-systems,references/urls}
\bibliographystyle{plain}

\end{document}